\documentclass[10pt,journal,compsoc]{IEEEtran}
\usepackage{amsmath}
\usepackage{amsthm}
\usepackage{mathrsfs}
\usepackage{bbm}
\usepackage{graphicx}
\usepackage{subfigure}
\usepackage{mathrsfs}
\usepackage{multirow}
\usepackage{balance}  
\usepackage[vlined,boxed,commentsnumbered, ruled, linesnumbered]{algorithm2e}
\usepackage{amsfonts,amssymb}
\usepackage{array}
\usepackage{setspace} 
\usepackage{xcolor}
\usepackage{booktabs}
\usepackage{hyperref}

\ifCLASSINFOpdf
\else
\fi
\begin{document}
%
\title{Generating Semantic Adversarial Examples via Feature Manipulation in Latent Space}
%
%
%
%

\author{Shuo~Wang,~\IEEEmembership{Member,~IEEE,}
		Surya~Nepal,~\IEEEmembership{Member,~IEEE,}
		Carsten~Rudolph,~\IEEEmembership{Member,~IEEE,}
		Marthie~Grobler,~\IEEEmembership{Member,~IEEE,}
		Shangyu~Chen,
  and~Tianle~Chen
\IEEEcompsocitemizethanks{\IEEEcompsocthanksitem Shuo Wang is with CSIRO's Data61 and Cybersecurity CRC, Australia.\protect\\
E-mail: shuo.wang@monash.edu
\IEEEcompsocthanksitem Carsten Rudolph and Tianle Chen are with Faculty of Information Technology at Monash University, Melbourne, Australia.
\IEEEcompsocthanksitem Surya Nepal and Marthie Grobler are with CSIRO's Data61 and Cybersecurity CRC, Australia.
\IEEEcompsocthanksitem Shangyu Chen is with University of Melbourne, Melbourne, Australia.
}}

\IEEEtitleabstractindextext{%
\begin{abstract}
\textcolor{black}{
The vulnerability of deep neural networks to adversarial attacks has been widely demonstrated (e.g., adversarial example). Traditional attacks perform unstructured pixel-wise perturbation to fool the classifier, which results in distinguishment from natural samples and lacks interpretability for human perceptions. 
Alternative approaches are to explore coarse-granularity structural perturbations in the latent space. However, such perturbations are hard to control due to the lack of interpretability and disentanglement. 
In this paper, we propose a practical adversarial attack by designing fine-granularity structural perturbation with respect to semantic meanings. Our proposed technique manipulates the semantic attributes of images via the disentangled latent codes. 
We design adversarial perturbation by manipulating a single or a combination of these latent codes and propose two unsupervised semantic manipulation approaches: vector-based disentangled representation and feature map-based disentangled representation, in terms of the complexity of the latent codes and smoothness of the reconstructed images. 
We conduct extensive experimental evaluations on real-world image data to demonstrate the power of our attacks for black-box classifiers. We further demonstrate the existence of a universal, image-agnostic semantic adversarial example.
}
\end{abstract}

}

\maketitle

\IEEEdisplaynontitleabstractindextext

%
\IEEEpeerreviewmaketitle

\IEEEraisesectionheading{\section{Introduction}\label{sec:introduction}}
\textcolor{black}{
The existence of adversarial examples causes security concerns about the reliability of deep neural networks (DNNs). In general, imperceptible perturbations that are hard for humans to recognize are intentionally added to the original inputs to cause a DNN to mislabel the perturbed inputs with high confidence \cite{papernot2016limitations,liu2016delving}. 
In terms of the adversarial investigation scope, we divide adversarial attacks into three categories, namely, nonstructural (pixel-level), semi-structural (layer-level), and structural (attribute-level). 
\newline
\textbf{Nonstructural level perturbations (pixel-level).} 
Present adversarial attacks typically involve manipulating artificially crafted, imperceptible perturbations that are nonstructural in the pixel-wise input space. 
As a consequence of these norm-bounded pixel perturbations, the adversarial examples are somewhat distinguishable from natural images. 
Multiple defense methods utilize this property to conduct adversarial training on norm-bounded pixel perturbations against small perturbations or to realign a perturbed image back into the space of a natural image by applying preprocessing filters or reconstructing the image. 
Moreover, such nonstructural perturbations are difficult to interpret from a human perspective. 
\newline
\textbf{Semi-structural level perturbations (layer level).} 
Existing studies also exploit a new class of adversarial examples with structural perturbations in order to provide some degree of interpretability for human perception. 
In particular, adversarial transformations are developed to create adversarial examples. Common transformations \cite{hosseini2018semantic,bhattad2019unrestricted,engstrom2018rotation}, such as rotation, clipping, or RGB conversion, are based on the shape bias property of human perception. 
However, the degree of granularity in such transformations is determined by the layer properties of the image, which are static, limited, and not always effective for all images. Therefore, we classify them as semi-structural level or layer level perturbations. 
\newline
\textbf{Structural level perturbations (attribute level)}
Additionally, some recent works have attempted to develop transformations in the latent space to obtain structural level adversarial perturbation, e.g., \cite{zhao2017generating}. 
However, these transformations fail to capture the correlation between the perturbation and attribute space that a human can understand. 
Random perturbations often lead to substantial differences in the input space, which are difficult to interpret. 
Towards fine granularity attribute-level perturbation, the interpretability and controllability of the latent transformation are crucial. 
}

Recent developments in variational autoencoder (VAE) \cite{larsen2015autoencoding}-based representation learning, e.g., $\beta$-VAE \cite{higgins2017beta}, Factor-VAE \cite{kim2018disentangling}, have resulted in significant advancements in the fields of disentanglement learning. These approaches provide an interpretable way to understand and control high dimensional data samples in terms of a low dimensional set of latent representations. Also known as disentanglement, this is a representation where a change in one dimension corresponds to a change in one factor of variation while being relatively invariant to changes in other factors \cite{bengio2013representation}. Consequently, it is feasible to conduct a new adversarial attack by manipulating disentangled latent representations (also called latent codes). Latent codes consist of a latent vector or a set of latent feature maps.

\textcolor{black}{
In this paper, we propose a systematic framework for creating structured perturbations in terms of attribute variation. This framework is designed to produce adversarial examples that appear to have a smooth and natural visual transition for human perception while misclassifying examples efficiently. Specifically, we leverage the disentangled generative model to capture the connection between the disentangled latent space and the semantic attributes in the pixel space, resulting in interpretability and controllability over the attribute modification (demonstration is given in Section~\ref{sec_demo}).
In general, images from a specific domain (e.g., handwritten numbers) share some common semantic attributes. Attributes that are irrelevant to a particular victim model (such as number classifiers) are termed theme-independent attributes, whereas the rest are theme-relevant attributes. For example, the font, font style (italic, oblique), and font size of handwritten numbers are theme-independent attributes of the number classifier, whereas the attributes that determine the identity of the numbers are theme-relevant. 
While it is difficult to separate theme-relevant attributes from these learned features, a few theme-independent attributes can be determined for a given classifier. 
Therefore, we assume that the semantic attributes can be factorized or disentangled in the latent space of the generative model, and theme-independent attributes can be derived within the disentangled latent space.  
Leveraging the disentangled latent codes, we design adversarial perturbation by manipulating a single or a combination of these latent codes relevant to the theme-independent attributes in an interpretable and structural manner. 
For example, we find a latent code that only controls a specific semantic attribute (such as the thickness) of handwritten digits. Then, we manipulate the attribute of a digit "7" by adding perturbations to the corresponding encoded latent codes until the classifier misidentifies the perturbed digit as a "1".
Taking face recognition as another example, the adversary may be able to perform an adversarial semantic attack by identifying the disentangled latent code and revealing only smiling facial expressions. This could fool a face classifier into making the wrong prediction by conducting linear interpolation on the latent feature map to change a normal face with a smiling facial expression. Ideally, the change of the adversarial example should be natural and smooth for humans' perception.  
}

\textcolor{black}{
In this work, we present generated adversaries to demonstrate the potential of the structural adversarial attack for black-box classifiers, with respect to semantic theme-independent attributes that humans can understand. 
The main contributions of this work are summarized as follows: 
\newline
(1) To obtain latent representation that can easily be manipulated in a human-perceptual manner, we propose a common feature variational autoencoder (CF-VAE), with better disentanglement performance. CF-VAE is trained to obtain bidirectional mapping between high-dimensional samples $x$ and low-dimensional latent representations $z$ (also known as latent codes), which are disentangled and have semantic meaning. To achieve a better disentanglement, the CF-VAE divides the latent codes into theme-independent codes and theme-relevant codes. In addition, a Total Correlation term is used to improve independence in the distribution of latent code $z$ to further improve disentanglement performance. 
\newline
(2) To balance the trade-off and improve reconstruction quality, we propose a generative adversarial network-based booster (GAN-B) to learn a more accurate projection of mapping latent code to the pixel-wise domain. This is necessary since the utility of the reconstructed sample by the decoder is generally limited and blurry as a result of the trade-off between reconstruction and disentanglement. The feature-wise similarity is used to transfer the representation from the pre-trained GAN (generative adversarial network) to the CF-VAE for reconstructing high-quality instances. Namely, the latent representation of a single layer $l$ from the discriminator of GAN (trained on clean instances) is used as the reconstructed error evaluation for CF-VAE.
\newline
(3) We design a perturbation method on the theme-independent latent codes to reconstruct the perturbed latent codes into an adversarial example that causes misclassification. 
To find the optimal perturbation added to the latent code, both from a visual perspective and from a latent space perspective, while ensuring the stealthiness of the malicious perturbation so that it can defeat latent detection-based defenses. The attack-effective value range for perturbation design that can effectively give rise to misclassification for a specific (or a combination) latent code is also recognized. 
\newline
(4) We enhance the semantic adversarial attack against images with complex semantic attributes, e.g., face images, by combining the CF-VAE with image-to-image translation. To obtain a better smoothness of perturbed images by reconstruction, we jointly train two CF-VAEs to obtain theme-independent latent codes, i.e., a set of feature maps, for semantic manipulation. 
\newline
We conduct extensive experimental evaluations on the real-world image data to demonstrate the power of our attack for black-box classifiers and present a comprehensive empirical analysis of the possible factors that affect the attack. Our experiments reveal potential security breaches where adversaries can control the semantic features to break a classifier and demonstrate that semantic-based perturbation is practical. We also empirically investigate the universal semantic adversarial examples also exist. 
}
The next section describes the preliminary concepts and the background on adversarial examples. Sections 3 and 4 explain the system design and our approach. Section 5 describes our experimental results. Section 6 discusses related work, and Section 7 concludes the work as a whole.


\section{Background}

\textbf{Generative Adversarial Networks.} 
The typical GAN architecture contains two neural networks: one generator neural network $G$, trained to generate a sample from a set of random numbers, and one discriminator $D$, trained to categorize data as real or fake. As the GAN networks are trained, $G$ learns to generate samples that can fool $D$. 
Let $p_z (z)$ be the input noise distribution of $G$ and $p_{data} (x)$ be the real data distribution, the purpose of GANs is to train $G$ and $D$ to play the following two-player minimax game with value function $V (G,D)$:
\begin{equation}
\small
\begin{aligned}
\min \limits_{G}\max \limits_{D}V(G,D)=E_{x }[logD(x)]+ \\ E_{z}[log(1-D(G(z)))]
\end{aligned} 
\end{equation}
GAN can be applied in varied unsupervised and semi-supervised learning tasks \cite{chen2016infogan,donahue2016adversarial,kumar2017improved,ledig2017photo,reed2016generative}. 

\textbf{Autoencoders and $\beta$-VAE.} 
Basically, an encoder $E$, parameterized by $q_{\phi }(z|x)$, is trained to convert high-dimensional data x into the latent representation bottleneck vector $z$ in latent space that follows a specific Gaussian distribution $p(z)  \sim N(0, 1)$. The decoder $p_{\theta}(x|z)$ is trained to reconstruct the latent vector $z$ to x. 
The training process of the autoencoders is to minimize the reconstruction error. 
The VAEs model \cite{larsen2015autoencoding} shares the same structure with the autoencoders but is based on an assumption that the latent variables follow some kind of distribution, such as Gaussian or uniform distribution. It uses variational inference for the learning of the latent variables. In VAEs the hypothesis is that the data is generated by a directed graphical model $p(x|z)$ and the encoder is to learn an approximation $q_{\phi} (z|x)$ to the posterior distribution $p_{\theta}(z|x)$ estimated by the decoder. 
The encoder and decoder are trained simultaneously based on the negative reconstruction error and the regularization term, i.e., Kullback-Leibler (KL) divergence between $q_{\phi }(z|x)$ and $p(z)$, by optimizing the variational lower bound:
\begin{equation}
\small
L(\theta ,\phi ;x) = KL(q_{\phi }(z|x)||p(z)) - \mathbf{E}_{q_{\phi }(z|x)}[log p_{\theta }(x|z)] 
\end{equation}
The left part of Equation (2) is the KL divergence regularization term to match the posterior of $z$ conditional on $x$, i.e., $ q_{\phi }(z|x)$, to a target distribution $ p(z)$, e.g., Gaussian distribution whose mean $\mu$ and diagonal covariance $\sum$ are the encoder output. 
The right part of Equation (2) denotes the reconstruction loss for a specific sample $x$. 


$\beta$-VAE \cite{higgins2017beta} is a modification of the VAE framework that introduces an adjustable hyperparameter $\beta$ to the original VAE objective: 
\begin{equation}
\small
\mathcal{L} = \mathbb{E}_{q_{\phi }}(log p_{\theta}(x|z))-\beta D_{KL}(q_{\phi }(z|x)|| p(z))
\end{equation}
\section{Problem Definition}
\subsection{Semantic Adversarial Example}
\textcolor{black}{
Unlike the notion of the adversarial example introduced in Section 2, the semantic adversarial example can be defined and formalized as follows. 
Let $D$ denote the set of examples in the sample space.
Let $\mathcal{F}(x)$ denote a pre-trained classifier, which outputs the final prediction label $y=\mathcal{F}(x)$. 
Let $S$ denote the set of underlying semantic features, interpreted as the representation of the data by human cognition. For example, $s_1 \in S$ is the color of skin and $s_2 \in S$ is the length of hair for a given human face image. 
We assume that observations $x \in D$ can be generated by combining $M$ primary disentangled latent factors $S_{o}= (s_1, \cdots, s_M), s_i \in S$. Besides, we assume that $ S_{c}= (s_1, \cdots, s_K), s_i \in S_o $ is the set of $K$ commonly shared theme-independent features with respect to the classification theme, e.g. the font size of handwritten digits for number classifier or facial express of face images for face recognition model. 
Unlike existing adversarial examples that directly add unstructured pixel-wise perturbation in the input space or random latent representation of z space, our adversary $A$ aims at perturbing a fraction of disentangled latent representations derived from a sample $x$ to generate structured perturbations in terms of  theme-independent semantic features. 
Such structural perturbation is named semantic feature manipulation, which is achieved by (1) learning a latent code that can reveal the commonly shared theme-independent features across classes, e.g., via VAE-based disentangled learning; (2) perturbing latent codes to reconstruct an adversarial example that can fool a targeted classifier in an interpretable and structural manner.
}

Let $E(x)$ be an encoder that can map $x$ to latent codes $\overrightarrow{z}$, namely $ E(x) \rightarrow \overrightarrow{z}=\{ z_1, \cdots, z_M\}$, in which $z_i$ reveals a specific semantic feature $s_i$ and $z_i$ follow the distribution $P_z$. $\overrightarrow{z}$ can be inversed back to $x^*$ with the help of a decoder $D(\overrightarrow{z}) \rightarrow x^*$.
The perturbation of latent codes is to add random noise $\Delta z$ to the values of the latent codes, assuming $z_i$ for simplicity. 
$\Delta z$ is from the same distribution $P_z$ as $z$.
The goal of $A$ is to design perturbation $\Delta z$ for $z_i$ to fool the pre-trained classifier $f$ such that $ \mathcal{F}(x^*) \neq y, \ D( E(x)=\{ z_1, \cdots , z_i +\Delta z, \cdots \} ) \rightarrow x^*$ while $\Delta z$ to be imperceptible and $x^*$ to be as similar as  $x$  (according  to  particular smoothness metrics).

The adversarial image $ x^*$ for $x$ can be designed by satisfying the three objectives given as follows, with respect to the reconstructed image based on the perturbed latent codes. 
\begin{equation}
\small
\begin{split}
x^*=D(E(x)=\{ z_1, \cdots , z_i +\Delta z^* , \cdots \} ),\\ s.t. \ \mathcal{F}(x^*) \neq y \ and \ ||\Delta z^*||\leq \epsilon,  \\  \Delta z^*  = \underset{}{argmin} ||\Delta z||,  \\ and \ similarity(x^*, x) \geq \theta_{SIM} 
\end{split}
\end{equation}

\textcolor{black}{
We also define the universal semantic adversarial example as follows. 
Let $D^{y}$ denote the set of examples with the same label $y$, the perturbation $\Delta z$, added in a specific latent code $z_i$ on a $x \in D^{y}$ to make misclassification, can also cause misclassification on a number of other $x' \in D^{y}$. The fraction of the affected instance is referred to as the fooling rate (FR). Namely, for the universal adversarial semantic adversarial example, not only the above equation should be satisfied but the fooling rate should also be more than a threshold $FR_{\Delta z^*}^{D^y} \geq \theta_{FR}$.
}

\subsection{Adversary Model}
We characterize an adversary according to its goals and knowledge regarding the learning model and training data. 
\subsubsection{Goals}
To launch an effective  semantic adversarial attack, the following goals of the adversary should be satisfied. 
\newline
(G1) Efficient perturbation to fool target classifier (term 1 in Eq. 6). A successful attack must have a high and consistent success rate (SR). The adversarial perturbation should be sufficiently reliable that a given perturbed instance is classified with high accuracy to the wrong label. A non-targeted misclassification attack is explored in this work. 
To ensure the consistent nature of the SR, the proportion of satisfactory perturbations that give rise to misclassification in high probability, within an interpolation value range $[low, high]$, a.k.a efficient ratio $ER$, should be more than a threshold $\theta_{ER}$.; 
\newline
(G2) Stealthy perturbation. It is desirable that the adversarial perturbation is stealthy so that it is hard to detect and should be visually imperceptible both in input space and latent space, even under the examination of a machine detector. Invisible perturbation should be added in the latent code so that the perturbed instance is close to the manifold of normal data in latent space (invisible/imperceptible, term 2 in Eq. 6), namely $\left | \Delta z^*_i \right |_p \leq \ \epsilon$. 
\newline
(G3) The perturbed instance has a high level of semantic similarity with the original instance, i.e., the change of the instance in the input space is considered as natural and smooth to the human eye (stealthiness, term 3 in Eq. 6), namely $Similarity(x^*,x) \geq \theta_{SIM}$; 
\newline
(G4) Perturbation should be as universal as possible, i.e., a single perturbation $\Delta  z^* $ can affect multiple instances, namely, the fooling rate should be more than $\theta_{FR}$. It is desirable that the adversarial perturbation is universal so that it is easy to design an effective perturbation for a large database without time and resource overheads.

\subsubsection{Knowledge}
\textcolor{black}{
In contrast to the minimal knowledge assumption that the adversary has known neither the training data nor the specifics of the learning model, we relax some of the assumptions related to the attacker's knowledge. Here, we assume that the adversary has no knowledge of the model architecture, i.e., black-box classifier, but has access to the training data, or either can know the type of training data the targeted classifier used, e.g., human face data or can obtain some very relevant data.
}
\section{Framework for Generating Semantic Adversarial Example}
In this paper, we propose a semantic adversarial attack scheme that is interpretable, smooth and universal. In this section, we describe the two approaches of our attack scheme: \textit{Vector-based Semantic Manipulation} and \textit{Feature Map-based Semantic Manipulation}, and its three main components, \textit{Learning Disentangled Representations}, \textit{GAN-Based Booster} and \textit{Invisible Latent Perturbation}.

\subsection{Attack Overview}
Our attack scheme aims to generate interpretable and universal semantic adversarial examples with invisible perturbation by manipulating some commonly shared or theme-independent features. 
Ideally, we can learn a disentangled latent code vector that can be used to reconstruct images and one factor of the vector affects one semantic feature only, via solely conducting naive common feature variational autoencoder (CF-VAE).
This may work well for simple images (e.g. handwriting digits), as the diversity and dimension of the feature are simple and most of the essential features can be captured via a 20-dimensional vector. 
However, the quality of the reconstructed image is significantly affected by the diversity and dimension of the feature retained by the latent codes, as well as the trade-offs between the disentanglement performance and the reconstruction quality. The performance on the complex images (e.g. face) is not as good as the simple images (e.g. handwriting digits) when only conducting onefold CF-VAE, since the feature dimension of the face is very large, and it is hard to leverage a low-dimensional latent code vector to generate a natural and smooth perturbed image.

Therefore, we apply two semantic manipulation approaches to generate adversarial instances in terms of the feasibility of semantic manipulation: \textit{Vector-based Semantic Manipulation via Onefold CF-VAE} and \textit{Feature Map-based Semantic Manipulation via Multiple CF-VAE}.
The advantage of the onefold approach is the unsupervised manner and that no corresponding image pairs are required. The advantage of the multiple approaches is the precise constraint on reconstituted images to achieve better quality. 

\begin{figure}[!htb]
    \centering
    \setlength{\abovecaptionskip}{-0.05cm}
    \setlength{\belowcaptionskip}{-0.2cm}
    \includegraphics[width=3.5in,height=2in]{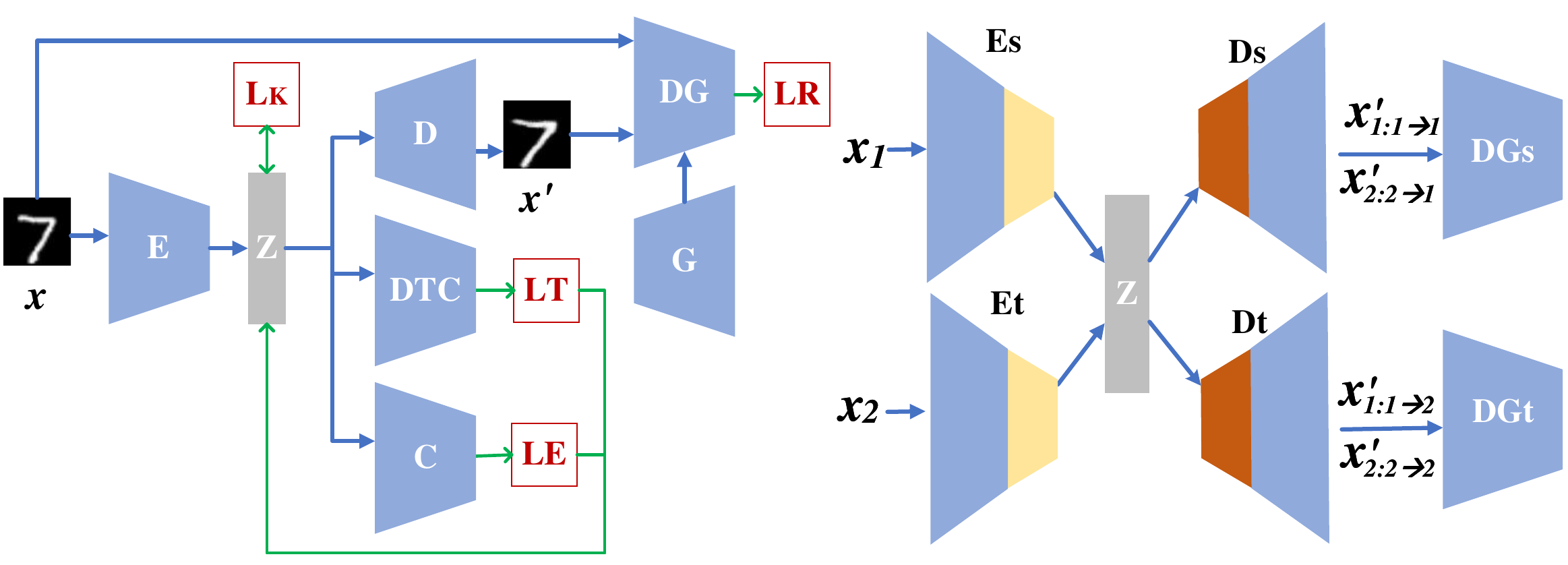}
    \caption{The vector-based semantic manipulation scheme via onefold CF-VAE (left) and feature map-based semantic manipulation scheme via multiple CF-VAEs (right). The loss function of the CF-VAE combines $L_K, \ L_R \ L_T$ and $L_E$. DTC (Discriminator for TC estimation) is trained to estimate the TC value, C is trained to strip class-unique features and DG (Discriminator of GAN) improves the reconstruction error evaluation.}
\end{figure}
\begin{figure}[!htb]
    \centering
    \setlength{\abovecaptionskip}{-0.05cm}
    \setlength{\belowcaptionskip}{-0.2cm}
    \includegraphics[width=3.5in,height=1.2in]{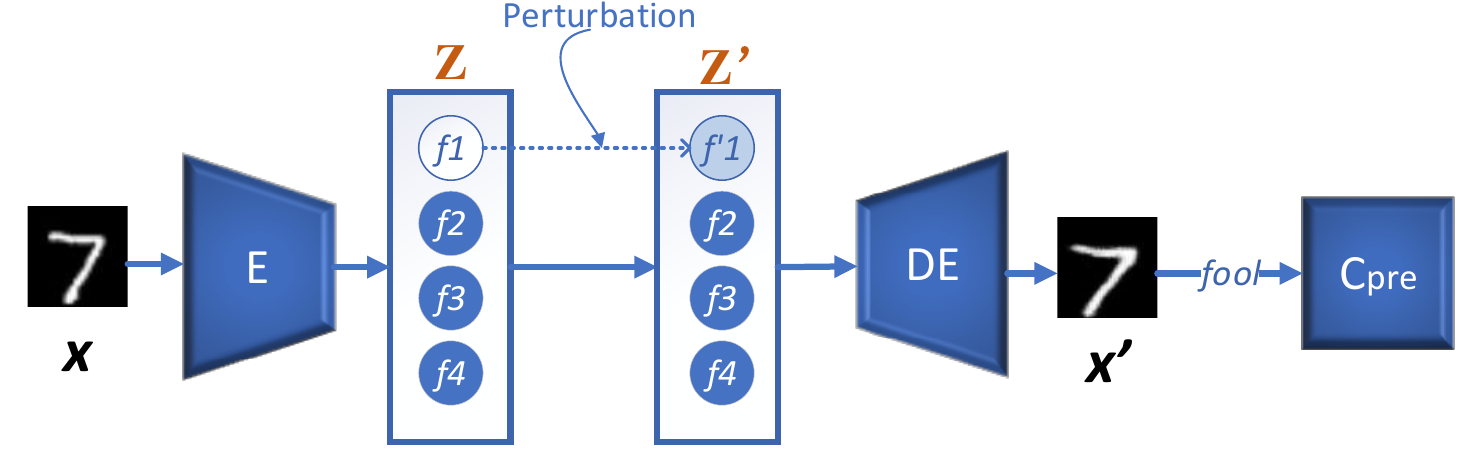}
    \caption{The semantic adversarial example perturbation scheme. Given an instance x, semantic adversarial attack generates adversaries by perturbing disentangled latent factor(s) and decoding the perturbation's latent bottleneck vector to fool the pre-trained classifier $C_pre$.}
\end{figure}
\textit{Vector-based Semantic Manipulation via Onefold CF-VAE}.
This can be implemented by a sole CF-VAE based representation learning, as shown in Fig.1 left, trained on clean training samples to obtain disentangled and independent latent representations that are easy to control and understand by humans.
To improve the disentanglement performance, a discriminator is trained for Total Correlation (TC) \cite{watanabe1960information} estimation (DTC) to improve the independence of latent representations for CF-VAE. In addition, a classifier $C$ is trained to filter class-unique information and only retain the common features to further improve the disentanglement performance. 
To improve the reconstruction quality, a generative adversarial network-based booster (GAN-B) is trained on clean training instances, which provides a discriminator $DG$ to improve the reconstruction error evaluation so that it enhances the quality of reconstructed instances. 
Finally, an invisible latent perturbation by stochastic searching is used to seek perturbation $z'_i=z_i + \Delta z$ for a specific disentangled latent factor $z_i$ (or a group), to achieve the adversarial goals, as shown in Fig. 2. 
Details of each component will be introduced in the following sections.


\textit{Feature Map-based Semantic Manipulation via multiple CF-VAE.}
For the complex images, we apply an unsupervised image-to-image translation, e.g. UNIT or MUNIT \cite{liu2017unsupervised,huang2018multimodal}, to learn the theme-independent latent codes for semantic manipulation. 
We learn the desired semantic attribute-conditional transition from well-labeled natural images, e.g. CelebA \cite{liu2015faceattributes} or RaFD dataset, for semantic manipulation. For instance, a set of smiley face images can be considered as a smile domain, the semantic transition from a natural face 
to smile can be conducted by learning a joint distribution of images in source and targeted (smile) domains. 
In this scenario, we assume that for any given pair of corresponding images $(x_s,x_t)$ in source and targeted domains, $X_s$ and $X_t$, can be mapped to the same latent representation in commonly shared latent codes $z$ \cite{liu2017unsupervised}. Here, the code is a set of feature maps, i.e. blocks of a feature matrix. Namely, both images can be recovered from this code, and we can compute this code from each of the two images. To achieve this goal, we jointly train two CF-VAEs on the source images $X_s$ (to attack) and target image $X_t$ (to change for, e.g., smile face images), as two domains, as shown in Fig. 1 right. Consequently, we can use $ D_t(z_s \sim q_s(z_s|x_s))$ to translate images from $X_s$ to $X_t$ by assembling a subset of the subnetworks. GAN-B can be also incorporated into the image-to-image transition to improve the quality of the reconstructed images. The latent code's iteratively stochastic searching is applied to generate adversarial examples by perturbing the shared latent codes (a set of the matrix) that control the desired semantic attributes.

\subsection{Vector-based Manipulation via onefold CF-VAE}
For uncomplicated  images, a low-dimensional latent code vector, learned by a onefold VAE-based disentanglement learning, can maintain most of the significant features of images. 
Then, it is desired to obtain better disentangled and theme-independent latent codes, which will be used to design smooth and natural adversarial examples via semantic manipulation-based perturbation. 
Consequently, we propose CF-VAE, derived from VAE. It is incorporated with a classifier on top of the latent code to achieve disentanglement in the latent codes $z$ that follow a fixed prior distribution while being unrelated to the label, i.e. to extract class irrelevant $z$. Furthermore, Total Correlation (TC) \cite{watanabe1960information} is applied to promote the latent factors to be more independent, to improve the disentanglement performance. 

\subsubsection{Disentangled theme-independent code learning}
To obtain theme-irrelevance for good disentanglement in $z$, we apply the irrelevance term $L_E$, estimated by a classifier parameterized by $\varsigma $ on the latent codes $z$, derived from the encoder $q_{\phi } (z|x)$, to classify the label of $z$. The adversarial learning approach, similar to \cite{makhzani2015adversarial}, is used to train the classifier. The adversarial classifier is added after z to distinguish its label, while the encoder tries to fool it, as demonstrated in Fig. 1 left. The objective of the classifier can be defined as cross-entropy loss:
\begin{equation}
\small
L_C=-E_{ q_{\phi } (z|x)}\sum_c I(c=y)log\ q_{\varsigma }(c|z)
\end{equation}

Here $I(c = y)$ is the indicator function, and $ q_{\varsigma }(c|z)$ is softmax probability output of the classifier. 
Simply, we assume the labels are distributed uniformly across all inputs, i.e. class probabilities $\pi = 1/C$. To peel the effect of labels 
and to extract class irrelevant z, the encoder $q_{\phi } (z|x)$  is simultaneously trained to fool the classifier with loss added by a cross entropy loss defined as follows:
\begin{equation}
\small
L_E=E_{ q_{\phi } (z|x)}\sum_c \frac{1}{C}log \ q_{\varsigma }(c|z)
\end{equation}
\subsubsection{Disentanglement improvement by inner-independence}
To obtain inner independence for good disentanglement in z, Total Correlation (TC) \cite{watanabe1960information} is used to encourage independence in the latent vector z, formally 
\begin{equation}
\small
TC(z) = KL(q(z)||\bar q(z)) = Eq(z)[log\frac{q(z)}{\bar q(z)}] 
\end{equation}

As TC is hard to obtain, the approximate tricks used in \cite{kim2018disentangling} is applied to estimate TC. Specifically, a discriminator $DTC$ is applied to output an estimate of the probability D(z) whose input is a sample from $q(z)$ rather than from $\bar q(z)$ to classify between samples from q(z) and $\bar q(z)$, thus learning to approximate the density ratio needed for estimating TC \cite{kim2018disentangling}. $ DTC $, parameterized by $\upsilon $, is trained with the other CF-VAE components jointly. Thus, the TC term is replaced by the discriminator-based approximation as follows:
\begin{equation}
\small
L_T=TC(z) \approx E_{q(z)}[log \frac{D(z)}{1-D(z)}]
\end{equation}
\subsubsection{Onefold CF-VAE Training}
The objective of CF-VAE is augmented with a TC \cite{watanabe1960information} term and $L_E$ to encourage independence in the latent factor distribution, given as follows:
\begin{equation}
\small
\begin{aligned}
\frac{1}{N}\sum_{N}^{i=1}[E_{ q_{\phi } (z|x^{(i)})}[log p_{\theta} (x^{(i)}|z)] - D_{KL}( q_{\phi } (z|x^{(i)})||p(z))] \\ - \gamma L_T -L_E
\end{aligned}
\end{equation}

This is also a lower bound on the marginal log likelihood $E_{p(x)}[log p(x)]$. 
The first part reveals the reconstruction error, denoted by $L_R$, evaluating whether the latent codes $z$ is informative enough to recover the original instance, e.g., $l_2$ loss between the original instance and the reconstructed instance. 
The second part is a regularization term, denoted by $L_K$, to push $ q_{\phi } (z|x)$ to match the prior distribution $p(z)$.
The third part is the TC term, denoted by $L_T$ (Equation 10), to measure the dependence for multiple random variables. 
The last part is the irrelevance term $L_E$ (Equation 8).

The parameter $\phi $ of encoder $ q_{\phi } (z|x)$ is then trained by $L_K$, $L_E$, $L_R$ and $L_T$ in terms of $-\nabla_{ \phi }(-L_{R} +\ L_{K} +   \gamma L_T+L_E) $, to let z be unrelated to the label, independent of each other and close to $N(0, I)$. 
The parameter $\varsigma $ of the adversarial classifier is updated in terms of $-\nabla_{\varsigma }(L_C) $.  
The parameter $\theta $ of decoder is updated in terms of $-\nabla_{\theta }(L_R) $.  
The parameter $ \upsilon $ of adversarial classifier is updated in terms of $-\nabla_{\upsilon }(L_T) $, i.e. $-\nabla_{\upsilon }\frac{1}{2|B|}[\sum_{i \in B}log (D_\upsilon (z^{(i)})+\sum_{i \in B'}log (1-D_\upsilon (permutedim(z'^{(i)}))] $ \cite{kim2018disentangling}.  Here, the permutedim function is to random permutate on a sample in the batch for each dimension of its $z$.

\subsection{GAN-Based Booster}
To address the trade-off between the reconstruction and disentanglement, we apply a GAN-based booster. 
Specifically, the GAN discriminator is used to learn a comprehensive similarity metric to discriminate real images from fake images. 
Inspired by \cite{gatys2015neural,larsen2015autoencoding}, the feature-wise metric expressed in the GAN discriminator, a.k.a. style error or content error is used as a more abstract reconstruction error $L_R$ for VAE to better measure the similarity between the original instance and the reconstructed image, aiming to improve the utility of the reconstructed instance. That is, we can use learned feature representations in the discriminator of a pre-trained GAN as the basis for the VAE reconstruction objective.

Specifically, let $x$ and $\hat{x}$ be the original image and the reconstructed image and $F^l(x)$ and $F^l(\hat{x})$ be their respective hidden feature representation of the $l^{th}$ layer of the discriminator, the representation loss to reveal reconstruction error is defined in Eq. 12 and 13. A Gaussian observation model is applied for $F^l(x)$ with mean $F^l(\hat{x})$ and identity covariance $ p(F^l(x)|z) = N(F^l(x)| F^l(\hat{x}), I) $. 
\begin{gather}
L^{F^l}_{R}= -E_{q(z|x)}[log \ p(F^l (x)|z))]\\
L^{content}_{R}(x,\hat{x},l)= \frac{1}{2}\sum_{i,j}(F^l_{i,j}-\hat{F}^l_{i,j})^2
\end{gather}

Simply, a GAN is trained in original clean instances. A single layer $l$ is then chosen from the discriminator network and used to obtain representation similarity according to $ F^l(x)$. 
The CF-VAE is then trained based on the representation error, leveraging the learned feature representations in the discriminator as the basis for the VAE reconstruction objective. 
Note that another variant of GAN, such as info-GAN \cite{chen2016infogan}, can be directly incorporated into our framework.

\subsection{Invisible Latent Perturbation by Stochastic Search}

In this section, we introduce an invisible perturbation approach to manipulate the latent codes $z_i \in z$ of an instance $x$, where $E(x) \rightarrow z $. We construct minimum perturbations $z^*_i= z_i+\Delta z^*$ to reconstruct adversarial samples $x^*=D( \hat{z}=\{ \cdots , z^*_i, \cdots \})$ that can fool a classifier into making wrong predictions over a number of instances with the common feature controlled by $z_i$. 

For simplicity, only one latent code factor is considered to be perturbed at one time. 
Let $\mathcal{F}$ be a neural network classifier to be attacked and $D$ be a set of training data samples for $\mathcal{F}$ or similar dataset. 
The perturbation mechanism iterative goes through each sample $x \in D$, and searches for the optimal perturbation in the neighborhood of $z_i \in E(x)=z$, within a searching range $[r,r+\Delta r]$ for $z_i$, to achieve misclassification and satisfy goals G1-4 in Section III-B. 

At each searching iteration, $N$ perturbations $\Delta z_i$ are randomly sampled within the current search range for evaluation. 
Effective Ratio (ER) is used as the indicator to evaluate the effectiveness of perturbation, defined as follows:
\begin{equation}
\small
ER=\frac{1}{N}\sum_{i=1}^{N}1_{\mathcal{F}(D(E(x)_{z_i}))\neq \mathcal{F}(x) } 
\end{equation}

To satisfy G1, misclassifications should be commonly achieved, i.e. ER exceeds a threshold $\theta_{ER}$. 
G2 is satisfied by the constraint of the searching range $||r+\Delta r|| \leq \epsilon$, where the updated perturbation is searched within the $L_p$ ball of radius $r$, $ \Delta z_i \sim (r,r+\Delta r)$.  
To satisfy G3, the structural similarity (SSIM) \cite{WangZ03} between the original input and the reconstructed instance should exceed a threshold of $\theta_{SIM}$.
The search range of perturbation $(r, r+ \Delta r )$ that is incrementally raised by $\Delta r$, until G1 to G3 are satisfied simultaneously. 
Among $\Delta z_i^*$, the one 
that has the closest distance to the original $ z_i$, is selected to reconstruct $x^*$ for $x_i$ as the adversarial example.

For the universal perturbation, the iterative searching is further evaluated in terms of universality on a set of data samples in the given class $y$, $D^y$, namely to evaluate whether a $ \Delta z_i^*$ from the previous steps satisfies the G4. 
Specifically, for each $\Delta z_i^*$ that satisfies G1-G3 simultaneously, its universality will be further evaluated using the universal indicator, fooling rate (FR), defined as follows:
\begin{equation}
\small
FR=\frac{1}{m}\sum_{i=1}^{m}1_{\mathcal{F}(D(E(x_m)_{z_i}))\neq \mathcal{F}(x_m)} 
\end{equation}

Here, the validation dataset is $X_{ \Delta z^*, z_i } = \{ D( E(x_1)_{z_i}  + \Delta z^*),\cdots, D( E(x_m)_{z_i}  + \Delta z^* \}$ modified from a batch of $m$ samples of $D^y$.
The searching is terminated when the empirical FR on the $X_{ \Delta z^*, z_i }$  exceeds the target threshold $\theta_{FR}$. 
The detailed algorithm is given in Algorithm 1.
\begin{algorithm}[!htb]
\small
    \KwIn{an instance $x$, search start value $r$ and upper bound $r_{max}$, $\epsilon$, target black-box classifier $f$, specific latent factor $z_i$, pre-trained encoder $E$ and decoder $DE$ from CF-VAE+GAN-B, threshold $\theta_{ER}, \theta_{FR},\theta_{SIM}$.}
    \KwOut{Universal perturbation $\Delta z^*$}
    S=$ \o $
    
    $ y = f(x)$, $z=E(x)$, r=0
    
    \While{$FR< \theta_{FR}$ or $ r+ \epsilon < r_{max} $}
    {
        $\{\Delta z\} \leftarrow $ sample N random noise within $[r, r+\epsilon]$
        
        count =0
        
        \For {$\Delta$ z in $\{\Delta z\}$}
        {
            $z'_i=z_i+\Delta z $, $\hat{x}=DE(\hat{z})$, $\hat{y}=f(\hat{x})$
            
            \If{$\hat{y}\neq y$}
            {
                count ++, S.add($\Delta z $), SIM=$similarity(\hat{x},x)$
            }
            $X_{new} \leftarrow $ sample M instances from $X$
            
            FR= $\frac{1}{M}\sum_{i=1}^{M}1_{\hat{x} \neq y }, x_i \in  X_{new}$
            
        }
        ER=count/N
        
        \If{ SIM$<\theta_{SIM}$ or ER$<\theta_{ER}$ }
        {
            $r=r+\epsilon$
        }
        \Else
        {
            Return $\Delta z^*=argmin_{\Delta z \in S}$
        }
    }
    \caption{Invisible Latent Perturbation Algorithm}
    \label{alg:one}
\end{algorithm}
\subsection{Feature Map-based Manipulation via multiple CF-VAE}
To achieve semantic manipulation via image-to-image transition for the manipulation of complex images, 
two encoders $E_s$ and $E_t$ of CF-VAE are applied for mapping images to the same latent codes, while two decoders $D_s$ and $D_t$ are used to map latent codes to images in two domains, respectively. Given corresponding images pair $(x_s,x_t)$ from the joint distribution, then $z = E_s (x_s) = E_t (x_t)$ and conversely $x_s = D_s (z)$ and $x_t = D_t(z)$. The attribute-conditional semantic manipulation can be achieved via learning the function $ x_t =F_ {s-t}(x_s) = D_t (E_s (x_s))$ to map from $X_s$ to $X_t$. Such shared codes assumption implies the cycle-consistency assumption \cite{zhu2017unpaired}, i.e. $ x_s = F_ {t-s}( F_ {s-t}(x_s))$. 
These two CF-VAE are implemented based on the shared-latent space assumption using a weight sharing constraint \cite{liu2017unsupervised} and cycle-consistency (CC) \cite{zhu2017unpaired}. 
The weight-sharing constraint is used to relate the two CF-VAEs.

Specifically, the connection weights of the last few layers in $E_s$ and $E_t$ are shared to extract high-level representations of the input images in the two domains. Likewise, the connection weights of the first few layers in $D_s$ and $D_t$ are shared to decode high-level representations for reconstructing the input images.
When a pair of corresponding images can be well mapped to the same latent code and this same latent code is decoded to a pair of corresponding images, image-to-image translation occurs, i.e. an image $x_s$ in $X_s$ is translated to an image in $X_t$. This can be conducted through applying $D_t(z_s \sim q_s(z_s|x_s))$. The $X_s \rightarrow X_t$ and $X_t \rightarrow X_s$ can be trained jointly. 
In addition, we enforce the CC constraint in this joint training to further regularize the ill-posed unsupervised image-to-image translation problem. 

The learning object is given as follows:
\begin{equation}
\small
\begin{aligned}
\underset{E_{s,t},D_{s,t}}{min}L_{VAE_s}(E_s,D_s)+L_{cc_s}(E_{s,t},D_{s,t})+L_{GAN_s}(E_t,D_s,DG_s)\\
+L_{VAE_t}(E_t,D_t)+L_{cc_t}(E_{s,t},D_{s,t})+L_{GAN_t}(E_s,D_t,Dis_t)
\end{aligned}
\end{equation}
The VAE objects are the same with Equation 11. 
The GAN objective functions are given by conditional GAN objective functions:
\begin{equation}
\small
\begin{aligned}
L_{GAN_s} (E_t,D_s,Dis_s) = \lambda_{g}E_{x_s \sim P_{X_s}}[log Dis_s(x_s)] \\+ \lambda_{g}E_{z_t \sim q_t}(z_t|x_t)[log(1-Dis_s(D_s(z_t)))]\\
L_{GAN_t} (E_s,D_t,Dis_t) = \lambda_{g}E_{x_t \sim P_{X_t}}[log Dis_t(x_t)] \\+ \lambda_{g}E_{z_s \sim q_s}(z_s|x_s)[log(1-Dis_t(D_t(z_s)))]
\end{aligned}
\end{equation}
GAN objective functions are applied to ensure that the translated images resemble images in the target domains, respectively, using $\lambda_{g}$ to control the impact of the GAN objective functions.  
The CC constraint is modeled via a VAE-like objective function, similar to \cite{liu2017unsupervised}. If we jointly train the GAN with the VAE, this can be conducted by solving a mini-max problem, as in \cite{larsen2015autoencoding}. 
Consequently, we can use $ D_t(z_s \sim q_s(z_s|x_s))$ to translate images from $X_s$ to $X_t$ by assembling a subset of the subnetworks.

A latent code iterative stochastic searching is then applied to generate adversarial examples, similar to Algorithm 2 
. Note that, as the dimension of the latent codes can be very large in this scenario, they are stored as a set of feature map (matrix), e.g., a set of $256$ feature matrix with $64*64$ elements. We treat each matrix as a latent factor, then $\Delta z$ can be selected as the same value for each element of the matrix and the searching range is a $L_{64*64}$ ball of radius $r$. 

\subsection{Attack Demonstration}
\label{sec_demo}
\begin{figure}[!htb]
    \centering
    \setlength{\abovecaptionskip}{-0.05cm}
    \setlength{\belowcaptionskip}{-0.2cm}
    \includegraphics[width=3.6in,height=2in]{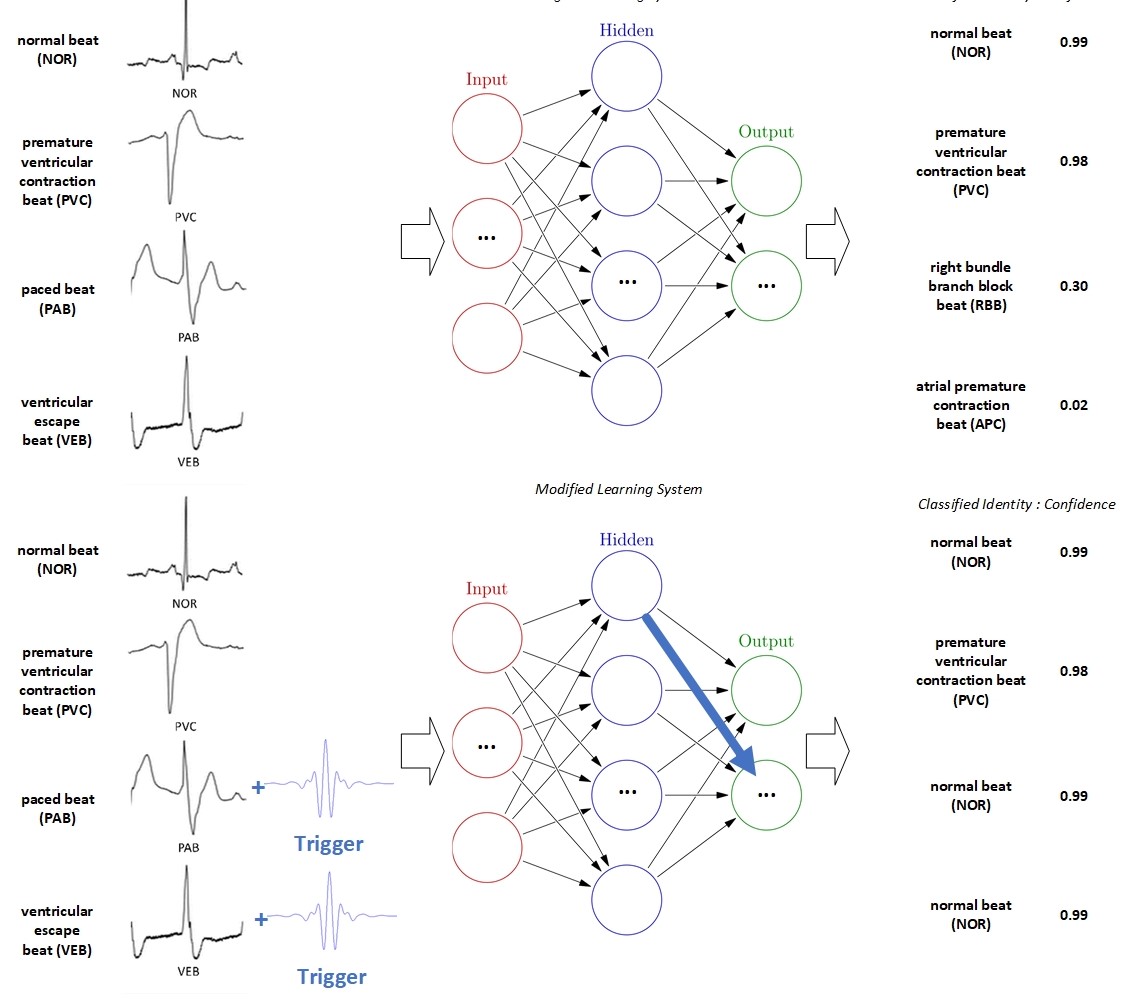}
    \caption{Semantic adversarial examples of MNIST. \rm{The figure to the left is the original digit "7" instance. Figures in the second part are perturbed instances that make misclassification (labels in front) by changing only one specific latent variable ($f_n$). Figures in the third part are the reconstructed instances, created by changing one latent variable on the original instance.}}
    \label{fig_digit}
\end{figure}

In this section, two types of semantic adversarial examples are demonstrated, using the vector and feature map based semantic manipulation, respectively.
A synthetic semantic adversarial example on the handwritten digits of MNIST \cite{lecun2010mnist} is shown in Fig.\ref{fig_digit}, which is used to demonstrate that an adversarial digit image can be designed by modifying only one latent code so that the change looks like something a person would write in order to fool both a classifier and human. 

\textcolor{black}{
Initially, an adversary trains a CF-VAE on a publicly-available handwritten digits dataset and applies the encoder of the CF-VAE to effectively map instance $x$ to its corresponding latent bottleneck vector $z$ composed of 20 disentangled latent codes. 
Then, the procedure to link latent code with semantic meaning is conducted as follow. 
\newline
\textit{\textbf{Linking latent code with semantic meaning.} }
As the latent codes learned by CF-VAE are already disentangled, we adopt a simple visualisation reconstruction to link the latent code with its underlying semantic meaning.
In particular, we randomly select one sample from each class, and then obtain its M-dimensional disentangled latent vector via the encoder of the well trained CF-VAE. 
For each latent code, we perform T times traversals on its value v between [-B,+B] with steps 2B/T while remaining the rest fixed. 
The traversed latent vector can be reconstructed to pixel space with the help of the decoder of the well training CF-VAE. 
Then, we visualize the T reconstructions using the T traversed latent vectors, as shown in each row in Fig. 3. 
By visualizing reconstructions of the traversed code, it is possible to infer its underlying semantic meanings. 
The identified semantic meaning could also be further confirmed by conducting visualisation reconstructions of other samples from the same class. 
With the knowledge of underlying semantic meanings, the adversary may select some latent codes that represent theme-independent semantic attributes as the victim latent codes to conduct perturbation. 
The perturbed latent vector can also be reconstructed to $x'$ using the decoder. There are 11 disentangled latent factors selected in Fig.\ref{fig_digit} in terms of semantic meaning, e.g. f7, f9, f13, f16, f17 respectively represent the degree of horizontal stroke crook, turning angle, thickness, rotation, and horizontal scale. Due to the disentanglement representation, each latent factor is exclusively responsible for the variation of a unique feature in the observed data. 
The perturbation on the factor will not significantly change the perceived quality of the original instance (e.g., it would look like a natural person's handwriting) and the reconstructed $x'$  should be similar to the original instance yet mislead classifiers' decision behavior on the input. 
}

It is easy to perturb each (or a group) latent variable to conduct the misclassification for the adversary in a human-cognition way. Taking the digit "7" in Fig.\ref{fig_digit} as an example, changing the value of latent variable f7 (degree of horizontal stroke crook) could fool LeNet \cite{lecun1998gradient} (test accuracy $99.1\%$) to change the label 7 to others. The latent codes that enable misclassification and retain the perceived quality of the reconstructed instance will be selected to conduct an attack in a more practical way, i.e., f4, f7, f10, f13. Furthermore, the adversary can add the same perturbation to the same latent factor for other "7" digits to evaluate the fooling ability and find a universal perturbation that causes misclassification generally for a large number of "7" digits.


\begin{figure}[!htb]
    \centering
    \setlength{\abovecaptionskip}{-0.05cm}
    \setlength{\belowcaptionskip}{-0.2cm}
    \includegraphics[width=3.6in,height=2in]{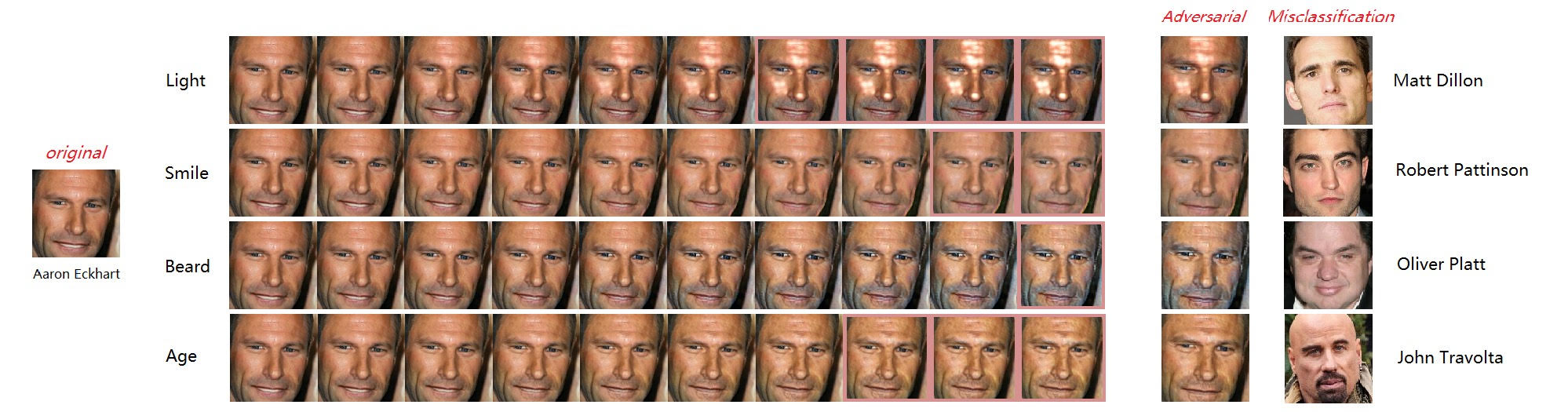}
    \caption{Semantic adversarial examples of CelebA. \rm{The figure on the left is the original face instance. Figures in the middle are the instances reconstructed by conducting linear interpolation with the same perturbation on each latent matrix of the original instance. The figures in the part on the right are perturbed instances that lead to misclassification by changing only one specific latent matrix that reveals a specific semantic features only.}}
    \label{fig_face}
\end{figure}

We also provide a synthetic semantic adversarial example demonstrated on the CelebA \cite{liu2015deep} database using the feature map based semantic manipulation, as shown in Fig.\ref{fig_face}.
The adversary jointly trains two CF-VAEs on the clean training samples or relevant source dataset (e.g., CelebA) and target dataset (e.g., only smile face in RaFD or CelebA), treated as the image-to-image transition. Then the input face image can be mapped into 256*64*64-dimensions latent codes, i.e. feature map, using the encoders of these two CF-VAEs. 
We find that some 64*64-dimensions matrices can reveal some disentangled semantic features. For example, we choose four matrices that respectively represent the degree of face light, smiling expression, beard and age, in Fig.\ref{fig_face}.
The adversary can choose a latent matrix that only impacts one specific semantic meaning, to conduct the semantic manipulation-based perturbation. Then the adversary will find an invisible and efficient perturbation for each selected latent factor, to cause the targeted classifier to misclassify the reconstructed instance. For example, the adversary can add perturbation to the second latent matrix to only change the smiling degree until the reconstructed face is misclassified. Furthermore, the adversary can add the same perturbation to the same latent matrix for other face images of this person to evaluate the fooling ability and find a universal perturbation that causes misclassification in general for this person.

\section{Experiments}
\subsection{Datasets}
In this section, we present the results from the empirical evaluations of the proposed approach over two types of benchmark datasets. 
(1) MNIST \cite{lecun2010mnist} consists of $28\times28$ grayscale handwritten digit images from 10 classes, i.e., digit 0-9 and has a training set of 55000 instances and a test set of 10000 instances. 
(2) CelebA (cropped version) \cite{liu2015deep} consists of 202,599 RGB $64\times 64 \times 3$ images of various celebrities, including 10,177 unique identities and 40 binary attribute annotations per image.





\subsection{Metrics} 
In this section, we evaluate the effectiveness of our proposed adversarial attack with experimental results on images. We rely on the following measures.
\subsubsection{Success Rate (SR)}
SR aims to measure the fraction of perturbed samples drawn from the test set being classified as the target class in the case of a targeted attack, changing to other arbitrary target class in the case of non-targeted attacks. A high SR shows an effective ability of the adversarial example to fool the classifier. SR is used to evaluate the vulnerability of trained classifiers to our semantic adversarial attack.
\subsubsection{Perturbation Stealthiness}
The stealthiness of perturbation represents the quality of the perturbation to generate an adversarial example. Although the perturbed samples with perturbation are likely visually imperceptible, it is preferable for the perturbation to be also imperceptible in latent space and evade input-preprocessing defenses. Therefore, three quantitative measurements are implemented to evaluate the stealthiness of semantic adversarial example via perturbation.
(1) Perceptual Hashing (pHash) Similarity: pHash \cite{liao2018backdoor} is used to present the fingerprint of an image based on its features. The pHash can measure the overall feature representation, instead of measuring the abrupt pixel change of an image. Images with similar features have similar pHash value. We can use the following equation to determine the similarity between the original image and the perturbed image and estimate how much the original image has been changed. 
\begin{equation}
S_{image}=(1-\frac{HD(pHash_{ori},pHash_{poi})}{64})\times 100\%
\end{equation}
where HD is the Hamming Distance, and 64 represent the binary length of the pHash score. 

(2) Structural Dissimilarity (SSIM). This is used to measure the internal feature similarity as a distance matrix evaluating the structural similarity of images \cite{WangZ03}. It is closer to the human sensitivities compared to $L_p$ distance. Therefore, using SSIM to evaluate the difference between two images satisfies human criterion more about negligible perturbation.
It evaluates the difference between two input images similar to human's criterion which is suit for the two image recognition applications in our experiments.

(3) Distance in the latent space (DLS). The difference in the input $x$ representation between the instance and its corresponding adversarial example, e.g., RMSE of the pixels, is not enough to quantify the semantic distance underlying them. Therefore, we use the distance in the latent space between them, i.e., $\Delta z =  || z^* - z'  ||$, to evaluate how much each instance is changed to achieve misclassification.

(4) The mean-squared error (MSE) between two images is used as the pixel-wise similarity measure. 

\textcolor{black}{
\subsection{Benchmark attacks and defenses}
\subsubsection{Benchmark attacks for efficiency}
We compare the attacking performance of our approach with state-of-the-art pixel level attacks and semantic level attacks.
For pixel level attacks, we implement FGSM, C\&W, DeepFool and PGD attacks. 
An existing similar adversarial attack on latent space is \cite{zhao2017generating}, in which the latent vector is not perturbed in an interpretable manner and without controllable mapping from latent space to pixel space. We simplify this type of attack as a general latent perturbation in our experiments, in which the latent vector is perturbed randomly. We compare our semantic attack with this general latent attack to demonstrate the controllability and feasibility of our attack. 
We also demonstrate the stealthiness (semantic similarity) of the perturbed instance for our attack, compared with general latent-based attacks. 
\subsubsection{Benchmark defenses for robustness}
We evaluated our attack when two strong defenses are present. 
One feasible defense is the input-based de-noising by a threshold of the pixel-wise reconstruction error. Therefore, we use the mean-squared error (MSE) between original and reconstructed instance via an autoencoder to reveal the robustness of our attack against such defenses; the varying threshold of MSE represents different levels of distinguishability of defenses. This can be considered as a typical input de-noising based defense, e.g., Magnet~\cite{meng2017magnet}.
\newline
Another stronger adaptive defense against our attack is to train the same autoencoder using relevant data and then find the normal value range for each disentangled feature factor. The adversary and protector are considered two parties of a game. The common parameter for defending and attacking in this game is the threshold to decide the normal variations of the latent representation.  The varying threshold of distance in the latent space (DLS) is used to reveal different levels of distinguishability of defenses. 
}
\subsection{Setup}
We use a Convolutional Neural Network for the encoder, a Deconvolutional Neural Network for the decoder and a Multi-Layer Perceptron (MLP) for the discriminator and classifier in CF-VAE for experiments on all data sets. We use [0,1] normalised data as targets for the mean of a Bernoulli distribution, using negative cross-entropy for log $p(x|z)$ and Adam optimiser with learning rate $10^{-4}$, $\beta_1 = 0.9$; $ \beta_2= 0.999$ for the VAE updates, as in \cite{higgins2017beta}. We also use Adam for the discriminator updates with $\beta_1 = 0.5$; $ \beta_2= 0.9$ and a learning rate tuned from $\{10^{-4},10^{-5}\}$. We use $10^{-4}$ for MNIST, and $10^{-5}$ for CelebA. The encoder outputs parameters for the mean and log-variance of Gaussian $q(z|x)$, and the decoder outputs logits for each entry of the image. We use the same six layer MLP discriminator with 1000 hidden units per layer and leaky ReLU (lReLU) nonlinearity, that outputs two logits in all CF-VAE experiments, as in \cite{kim2018disentangling}.
We train for $3 \times 10^{5}$ iterations on MNIST and $ 10^{6}$ iterations on CelebA. We use a batch size of 200 for all data sets. The default hurdle values to recognize satisfactory adversarial example are $[\theta_{FR}=0.2,\theta_{SSIM}=52,\theta_{ER}=0.1, \epsilon=0.1]$.

\subsection{Evaluations} 
In this section, we present and explain the high vulnerability of DNN classifiers to our semantic adversarial attack under various settings and demonstrate the perturbation stealthiness and universality as well. 
\subsubsection{Evaluation Attacking Performance Under Various Settings}
\textcolor{black}{
To demonstrate performance of proposed adversarial attack, we first compare such attack with other types of attacks, namely 
\textit{\textbf{non-structural}} (input space perturbation attack, e.g., projected gradient descent (PGD) \cite{madry2017towards}, DeepFool \cite{moosavi2016deepfool}, and Carlini Wagner Attack \cite{carlini2017towards}), 
\textit{\textbf{latent}} (perturbation is conducted generally into latent representations derived from autoencoder or inverter, e.g., natural adversarial example \cite{zhao2017generating}), \textit{\textbf{semantic}} (perturbation is conducted into disentangled latent representations derived from CF-VAE) and \textit{\textbf{boosted semantic}} (perturbation is conducted into disentangled latent representations derived from CF-VAE+GAN-B) settings respectively. 
We apply our adversarial attack to some state-of-the-art black-box classifiers for images, i.e. two-layer LeNet \cite{lecun1999object} convolutional neural network classifier for MNIST (with $99.1\%$ accuracy), 50-layer ResNet \cite{he2016deep} classifier for CelebA face ( $81.3\%$ accuracy for identification classifier). 
Beyond the success rate, we also demonstrate the structural property of perturbation between existing latent perturbation and our proposed semantic structural perturbation in Section 5.5.2. 
}

\textcolor{black}{
\textbf{Comparison with benchmarks}. We first report the success rate for attacks under \textit{\textbf{non-structural}} settings (input space perturbation attack, e.g., PGD, DeepFool, and CW), \textit{\textbf{latent}} settings ( latent perturbation, e.g., natural adversarial example \cite{zhao2017generating}), and \textit{\textbf{semantic}} settings (ours with and without boosted strategy). For each perturbation procedure and its corresponding setting, we evaluate the adversarial attack with each selected class. The attack SR on the test set is averaged among classes. 
The SR values for input space adversarial attacks are high (98.5\%, 89.8\% and 99.2\% for PGD, DeepFool and CW attacks on MNIST, respectively). SR of latent level attack can also achieve 98.5\%. However, our semantic adversarial attack could reach 100\% SR for both MNIST and CelebA with suitable ER hurdle (e.g., 10\%). 
Moreover, we find that our attacks can achieve 100\% success rates on samples where other attacks have failed. Our approach is demonstrated to be capable of performing an efficient misclassification attack on these missed samples.  
In addition, it is interesting to note that for the input space adversarial attacks require a high computational overhead to iteratively optimize the pixel perturbation using backpropagation on gradients. As models become more complex, the cost increases. In contrast, the proposed attack only requires feedforward computation on the pre-trained decoder and a black-box query to the victim classifier to discover the latent perturbation, thus avoiding the computation overhead problem. 
}

\textbf{Effect of Parameters}. In this section, we evaluate the effect of changing hyperparameters of the scheme on the results of images. Three scenarios are evaluated, e.g., the SR performance for \textbf{ordinary latent} (legend latent), general SR performance for \textbf{semantic} (legend semantic, in this case, we define a successful adversarial example when misclassification is achieved by perturbing any latent factor), and factor-specific SR performance for \textbf{semantic} (legend factor-n, in this case, we define a successful adversarial example when misclassification is achieved by only perturbing a specific latent factor).
An adversarial example candidate is decided as a satisfactory (or successful) adversarial example when the perturbation added to the latent factor(s) within a value range can achieve misclassification, while the efficient ratio ER and its structural similarity SSIM (or other similarity evaluation) are both more than thresholds $ \theta_{ER}$ and $\theta_{SIM}$ respectively. 

Therefore, we evaluate the SR change when varying the hurdle for ER and similarity evaluations (e.g., structural similarity SSIM, distance in latent space DLS and pHash similarity hurdles) respectively. We also investigate how the perturbation value range distribution over satisfactory adversarial examples in terms of the SR. 

As shown in the first subfigure of Fig. 5, the attack SR generally decreases when we increase the hurdle of ER to decide an adversarial example. In comparison, we can achieve a generally more decent attack performance with a high SR (more than $80\%$ for factor-specific evaluation even $100\%$ for general evaluation when ER hurdle is set as $10\% $) compared with an ordinary latent attack ($0\%$ SR when ER hurdle is set as $10\%$) as ER hurdle increases. Note that the SR will tend to be $0\%$ when the hurdle of ER is set more than a relatively large value ($50\%$ for the semantic latent attack while $5\%$ for ordinary latent attack).

The second to fourth subfigures of Fig. 5 describes the impacts of the similarity hurdle on the SR change. Generally, a great SSIM similarity or a small pHash and DLS similarity values reveal a good quality of the adversarial examples. When we raise the hurdle of SSIM evaluation or reduce the hurdle of pHash and DLS evaluations to decide an adversarial example, the attack SR generally declines. Namely, when we improve the desired quality criteria of adversarial examples, the number of satisfactory adversarial examples will decrease (it is harder to find satisfactory adversarial examples). Again, our attack can achieve an overall better attack performance as the criterion of similarity for satisfactory adversarial examples increases, compared with the ordinary latent attack.

We also demonstrate the impact of choosing various perturbation value ranges on the SR, namely how the perturbations of adversarial examples distribute over satisfactory adversarial examples. In order to conduct the semantic adversarial attack, an array of perturbations is added to various latent factors by varying the value range of factors exponentially from -10 to 10. As shown in the last subfigure of Fig. 5, most adversarial examples candidates are defined as satisfactory adversarial examples using the perturbation within a small perturbation range value for ordinary latent attacks, while the perturbation values that achieve satisfactory semantic adversarial examples are more evenly distributed. The reason is that the goal of ordinary latent attacks is to find the minimum perturbations that can cause misclassification only, no matter how the adversarial example differs from the original one in terms of some similarity metrics. However, our semantic attack aims to find appropriate perturbations that can cause misclassification via controlling specific feature(s), in which many determinants can affect the value range locating, e.g., the stealthiness in both the latent and input space. On the other hand, such range value distribution figures can provide a precise "handbook" for adversarial to design effective perturbation of adversarial attack, which can be experimentally obtained using a similar dataset as the targeted data.

\begin{figure}[!htb]
    \centering
    \setlength{\abovecaptionskip}{-0.05cm}
    \setlength{\belowcaptionskip}{-0.2cm}
    \includegraphics[width=3.6in,height=4.1in]{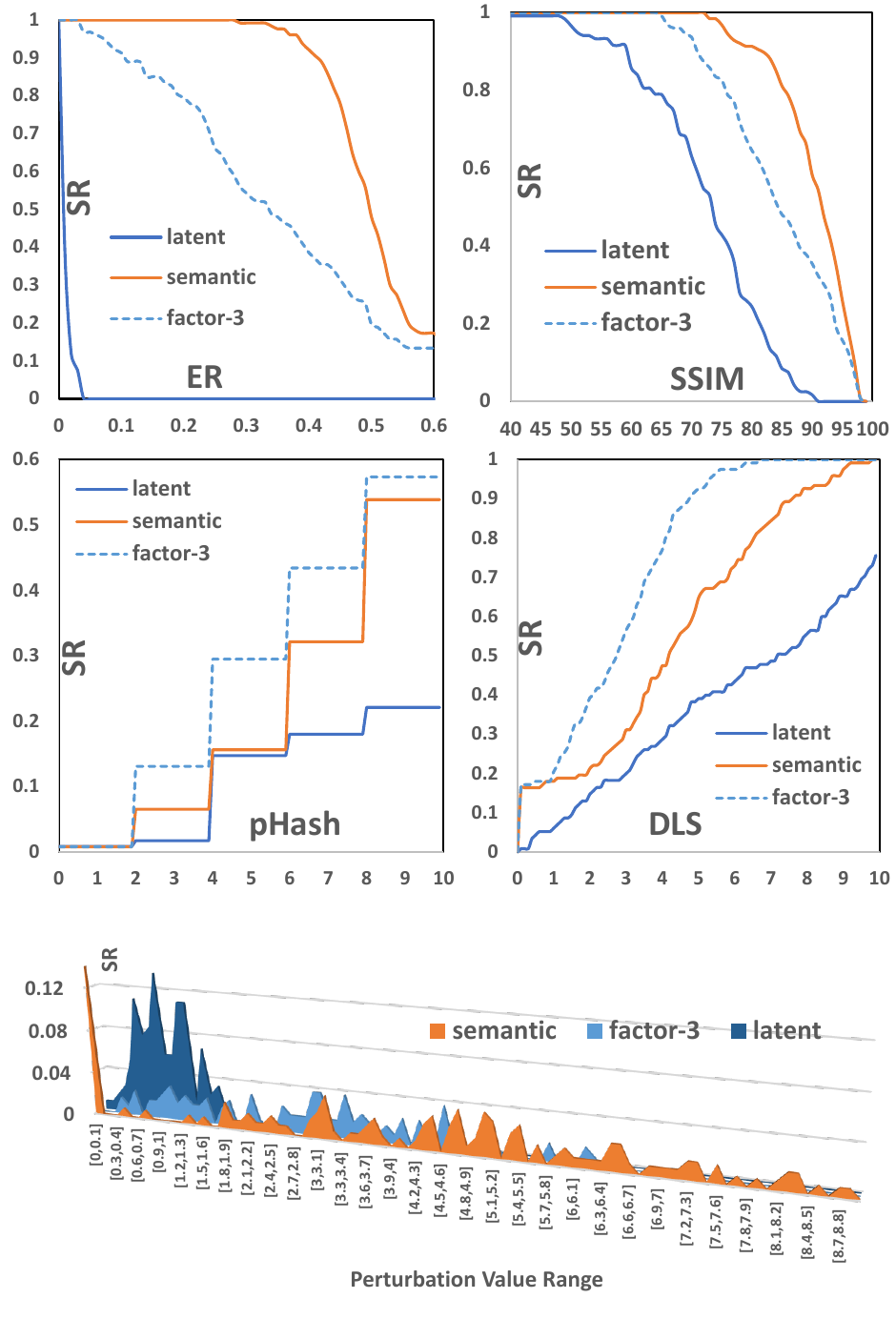}
    \caption{Success rate evaluation on MNIST. The first two rows reveal the SR change when varying hurdle of efficient ratio ER, similarity evaluation metric, e.g., structural similarity SSIM, pHash similarity and distance in latent space similarity.  The last figure demonstrates the success rate achieved on various perturbation range value.}
\end{figure}

\textbf{Effect of Selected Latent Factors}. Next, we will demonstrate the impact of selected latent factors on the SR in the following parts in detail. We evaluate the SR change when varying the hurdle for ER and similarity evaluations (e.g., structural similarity SSIM, distance in latent space DLS and pHash similarity hurdles) respectively, perturbing only one latent factor for each attack round. We also investigate how the perturbations of adversarial examples distribute over satisfactory adversarial examples when varying the latent factor to be perturbed.

As shown in Fig. 6, when we change the hurdle of ER, SSIM similarity, pHash, and DLS similarity values to decide an adversarial example, all factors have the same change tendency of attack SR. However, these detailed amplitudes of variation are different. Various factors show better performances in terms of SR at different hurdle criteria as well as different hurdle value ranges. Consequently, based on the experimental analysis in Fig. 5 and Fig. 6, the attacker can choose one appropriate factor to conduct the attack and choose the efficient value range for the selected features in terms of attack SR. 
We demonstrated the robustness of our attack against strong defenses in Fig. 5 and Fig. 6. The varying threshold of distance in the latent space (DLS) is used to reveal different levels of distinguishability of defenses. 

\begin{figure}[!htb]
    \centering
    \small
    \setlength{\abovecaptionskip}{-0.05cm}
    \setlength{\belowcaptionskip}{-0.2cm}
    \includegraphics[width=3.6in,height=4in]{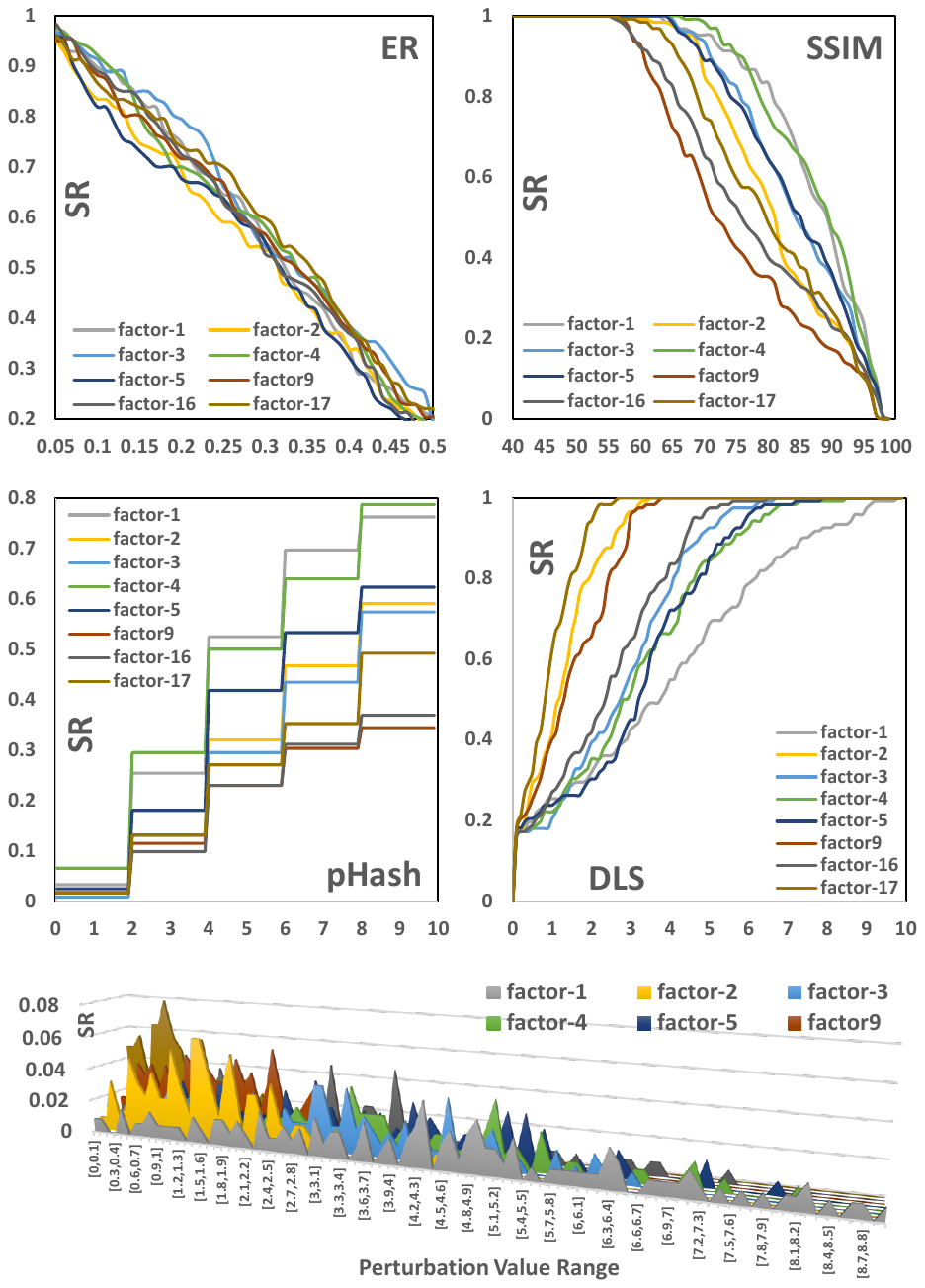}
    \caption{Success rate evaluation when perturbing different latent factor. The first two rows reveal the success rate change when varying hurdle of efficient ratio ER, similarity evaluation metric, e.g., structural similarity SSIM, pHash similarity and distance in latent space similarity.  The last figure demonstrates the success rate achieved on various perturbation range value.}
\end{figure}

\subsubsection{Evaluation of Structural Property (Stealthiness) and Robustness against Defenses}
\textcolor{black}{
We demonstrate the structural property of perturbation between existing latent perturbation and our proposed semantic structural perturbation in this section. 
We compare input, ordinary, semantic and boosted perturbation generated for the five class instances. For each class, we select 100 images from class $c$ in the testing set to implement these four perturbation mechanisms, then the average pHash, SSIM, MSE, and DLS similarity metrics are measured respectively. pHash and MSE are used to evaluate the pixel-wise similarity, DLS measures the latent similarity and SSIM estimates the similarity from the view of human cognition.
The results averaged among these metrics are summarized in Table I. 
}
\begin{table}[!htb]
\scriptsize
    \centering
    \caption{Evaluation of Perturbation Stealthiness}
    \label{tab:my-table}
    \begin{tabular}{|l|l|l|l|l|}
        \hline
        \multicolumn{2}{|l|}{Stealthiness} & latent & semantic & boosted \\ \hline
        \multirow{4}{*}{MNIST}    & pHash  & 9.42   & 9.14     & 7.82    \\ \cline{2-5} 
        & MSE    & 0.03   & 0.02     & 0.01    \\ \cline{2-5} 
        & SSIM   & 77.20\%  & 81.85\%     & 84.32\%    \\ \cline{2-5} 
        & DLS    & 7.01   & 2.23     & 1.16    \\ \hline
        \multirow{4}{*}{CelebA}   & pHash  & 39.53  & 18.41    & 16.21   \\ \cline{2-5} 
        & MSE    & 0.06   & 0.05     & 0.04    \\ \cline{2-5} 
        & SSIM   & 17.45\%   & 30.86\%     & 32.34\%    \\ \cline{2-5} 
        & DLS    & 8.32  & 5.97    & 5.24   \\ \hline
    \end{tabular}%
\end{table}

\textcolor{black}{
We can see that our $semantic$ has higher SSIM similarity scores while smaller pHash, MSE and DLS based dissimilarity scores (e.g., $81.5\%$, $9.14$,  $0.02$ and $2.23$ over MNIST), compared with the ordinary latent attack (e.g., $77.2\%$, $9.42$,  $0.03$ and $7.01$ over MNIST). The results reveal that our attack can generate adversarial examples with better quality in both input space and latent space to fool a time-limited human: the stealthiness of the perturbation is better ensured by our attack. 
We also find that the image with a $boosted$ strategy has a higher SSIM and smaller pHash, MSE and DLS dissimilarity values than that of the $semantic$ perturbation. This shows that the GAN-B improves the reconstruction quality in both input space and latent space. Namely, GAN-B can further enhance the perturbation of our attack. 
Similarly, the evaluations on the CelebA face data confirm these results again. 
We can see that $semantic$ has higher SSIM similarity scores while smaller pHash, MSE and DLS based dissimilarity scores (e.g., $30.86\%$, $8.14$,  $0.05$ and $18.41$ over CelebA), compared with the ordinary latent attack (e.g., $17.45\%$, $8.32$,  $0.06$ and $39.53$ over CelebA). The image with the $boosted$ strategy has a higher SSIM and smaller pHash, MSE and DLS dissimilarity values than that of the $semantic$ perturbation. 
Note that the similarity scores for face data are worse than hand-written digits. The reason is that the latent features of the face are more complicated than digits, which requires more well-labeled data and deeper neural networks to obtain better disentangled and reconstruction performance.
}
\begin{figure}[!htb]
    \centering
    \setlength{\abovecaptionskip}{-0.05cm}
    \setlength{\belowcaptionskip}{-0.2cm}
    \includegraphics[width=3.5in,height=0.8in]{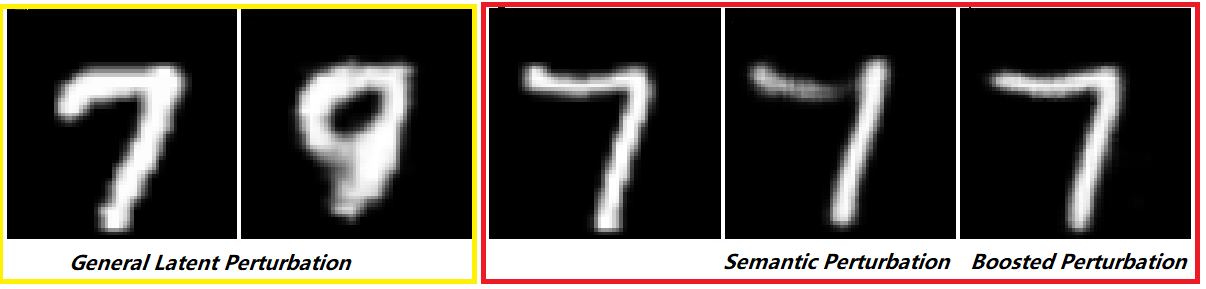}
    \caption{\textcolor{black}{Demonstration of stealthiness for our attack over MNIST. An ordinary latent attack is demonstrated in the yellow box (left is the clean instance and right is the perturbed instance that causes misclassification). A semantic attack is demonstrated in the red box (left is the clean instance, middle is a perturbed instance using semantic attack and right is the perturbed instance using boosted semantic attack).}}
\end{figure}

\textcolor{black}{
\textbf{Robustness against defenses. }
The MSE between the original and reconstructed instances using a well-trained autoencoder is used to identify malicious input. 
Various thresholds of MSE represent different levels of defense distinguishability. 
Under the threshold of 0.06 MSE for MNIST, more than 99\% of pixel-level adversarial examples can be identified, whereas less than 1\% of adversarial examples derived from our attack can be identified. 
MSE threshold value 0.1 for CelebA allows for the identification of more than 90\% of pixel-level adversarial examples, while less than 1\% of adversarial examples derived from our attack are identifiable. 
\newline
We also consider a stronger adaptive defense against our attacks. 
An effective defense is assumed to enable the training of an autoencoder using similar data to that used by attacks, allowing the normal variations of an encoded latent representation to be identified and used as a method to detect malicious data. 
Different thresholds of distance in the latent space (DLS) are used to identify the different levels of defense differentiation. 
According to our findings, there is a trade-off between detection accuracy and false-positive rate. The higher detection accuracy on adversarial examples, which we derive from latent perturbation and our semantic perturbation, comes at the expense of the proportion of incorrectly identified normal examples. 
Nonetheless, it is still possible to detect over 50\% of adversarial examples using a proper DLS threshold (such as 7 for MNIST and 9 for CelebA) for ordinary latent perturbation. Comparatively, the detection accuracy for our attacks is less than 5\%. The reason is that random perturbations on entangled latent representations may result in out-of-distribution samples. By contrast, perturbing well-disentangled latent codes can prevent such drift. In other words, after perturbing the disentangled latent representation, the reconstructed samples are more likely to represent the same distribution as the clean population. In addition, the false-positive rate rises further when considering well-structured adversarial examples that are almost identical to the clean ones. 
\newline
In addition, we also demonstrate the visual stealthiness of our attack. 
In Fig. 7, we compare the reversed instance from the perturbed latent variables for both ordinary latent attack and our semantic attack, taking produced adversarial examples for a single digit "7" as an example. 
The reversed instance by the ordinary latent attack is shown in the yellow box. We can see that such an attack might lead to misclassification by adding random noise into the latent variables, however, cannot control what the reversed instance looks like. In comparison, our semantic attack causes misclassification by perturbing a latent factor that only affects one feature, thereby controlling what the reversed instance looks like. As demonstrated in the red box, the perturbed instance using the semantic attack can achieve misclassification by only perturbing the latent factor that affects the stroke of the digit, while the quality of the reversed instance is further enhanced by using the boosted strategy.
We compare our semantic attack with this general latent attack in Fig. 5, 8, 10 and Table I, II to demonstrate the controllability and feasibility of our attack. 
}
\subsubsection{Evaluation of Universal Perturbation (Transferability) and Robustness against Defenses}
\textcolor{black}{
If we find a misclassification perturbation for an instance by adding noise to a specific latent factor, for example perturbing the factor that affects the stroke of digit "7", it is natural to consider whether the same perturbation, added to perturb the same latent factor of a different digit "7" instance, could give rise to misclassification.
Therefore, in this section, we investigate whether we can demonstrate the universality of our semantic adversarial examples across instances with the same class type. 
}

\textcolor{black}{
The goal is to evaluate whether the same perturbation on the same selected latent factor, leading to misclassification for a given classifier over a specific input $x$, can cause other instances from the same class with $x$ to be misclassified and how many and how to be impacted by hurdle parameters.
Specifically, we evaluate the universality of perturbation against two-layer LeNet classifiers on MNIST in terms of fooling rate FR. We select five class instances to obtain misclassification perturbation by only perturbating one latent factor $f_i$, which is composed of 500 images from the training set, i.e., on average 100 images per class, named construction set. The universal perturbation for each class on $f_i$ is the average misclassification perturbation on all 100 instances with the same class, named testing set. For each universal perturbation for $f_i$, we test the fooling rate over a validation set that consists of randomly selected 100 images with the same class type. Four latent factors are selected to be evaluated using the approach in Algorithm 1. 
}
\begin{figure}[!htb]
    \centering
    \setlength{\abovecaptionskip}{-0.05cm}
    \setlength{\belowcaptionskip}{-0.2cm}
    \includegraphics[width=3.6in,height=1.5in]{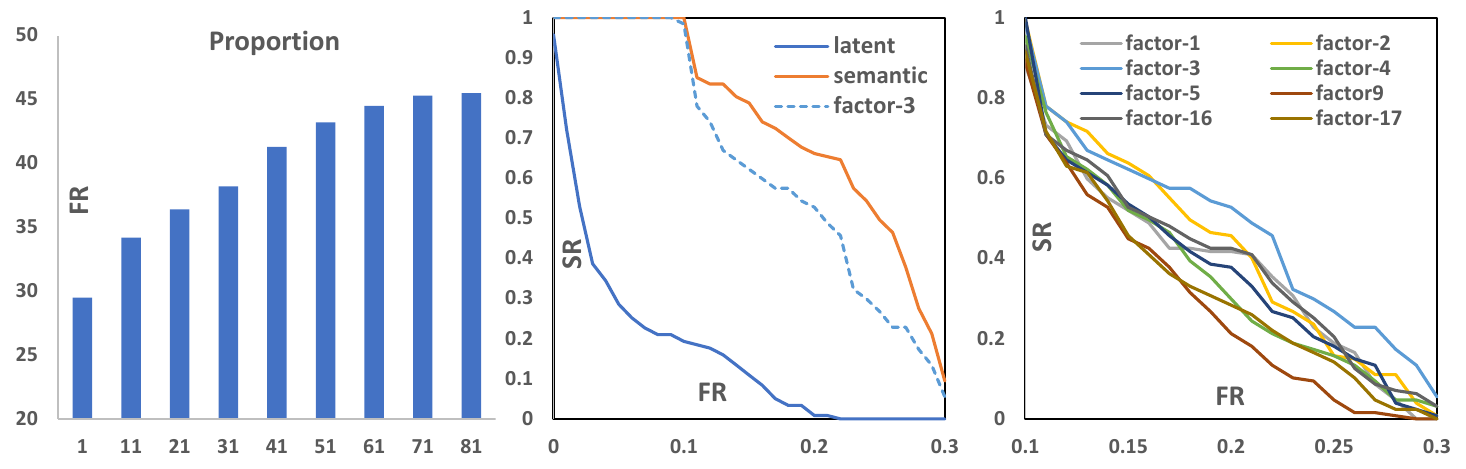}
    \caption{Fooling rate evaluation on MNIST. The success rate SR represents the proportion of adversarial example with corresponding fooling power in terms of the fooling rate at x axis.}
\end{figure}

\textcolor{black}{
The first subfigure of Fig. 8 shows the fooling rates obtained on the validation set when varying the proportion of the construction set to generate universal perturbation. The fooling rate increases as the size of the proportion goes up and then tends to be stable. 
An interesting result is that by using an arbitrary instance of the construction set to generate universal perturbation based on our semantic attack, we can fool approximately $30\%$ of the images on the validation set. 
\newline
Next, we use the one-instance-based universal perturbation to evaluate the fooling rate, to reveal the fooling power of the instance in the construction set. The last two subfigures of Fig. 8 reveal the distribution of adversarial examples in terms of fooling rate. 
Overall, the universal perturbation derived from any one instance of the construction set achieves very considerable fooling rates ($10-20\%$) on the validation set using our semantic latent attack. For example, more than $70\%$ of one-instance-based universal perturbation derived from our attack scheme can fool $20\%$ of clean images in the validation set.
The fooling power of universal perturbation, derived from any one instance of the construction set using an ordinary latent attack, is generally below $10\%$ (more than $80\%$ one-instance-based universal perturbation). 
Consequently, these demonstrations reveal that our attack has remarkable universal perturbation power over unseen data points, which can be computed on a small set of training instances, even arbitrary one instance. 
\newline
We further investigate the performance of the one-instance-based universal perturbation in terms of the fooling rate when varying parameter settings.
As shown in the first graph in Fig. 9, we evaluate the fooling rate change when varying the hurdle for ER and similarity evaluations (e.g., structural similarity SSIM and distance in latent space DLS hurdles) respectively when figuring out perturbation over the instances in the construction set using our semantic attack. As we change the hurdle of ER, SSIM similarity and DLS similarity criteria to decide an adversarial example, the fooling rate represents different trends of change. As we increase the hurdle of efficient ratio ER, the fooling rate vibrates a lot. However, it is feasible to choose an appropriate ER hurdle that can achieve a high fooling rate. When we raise the requirement for the similarity criteria of the adversarial examples, the fooling rate will decline. 
We also evaluate the impact of the selected latent factor on the fooling rate when varying parameters. Various factors show good performances in terms of fooling rate at different hurdle criteria.
}
\begin{figure}[!htb]
    \centering
    \setlength{\abovecaptionskip}{-0.05cm}
    \setlength{\belowcaptionskip}{-0.2cm}
    \includegraphics[width=3.5in]{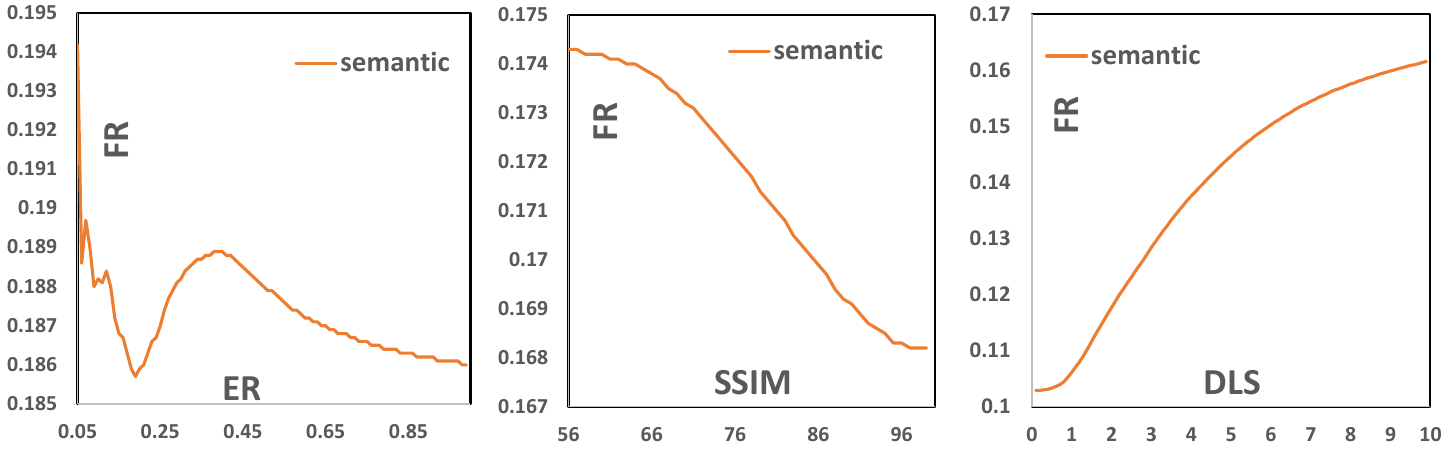}
    \caption{Fooling rate evaluation on MNIST with varying hurdle criteria to generate adversarial examples. 
    }
\end{figure}

\vspace{-0.4cm}
\begin{table}[!htb]

    \centering
    \scriptsize
    \caption{Evaluation of Stealthiness for Universal Perturbation}
    \label{tab:my-table}
    \begin{tabular}{|l|l|l|l|l|}
        \hline
        \multicolumn{2}{|l|}{Stealthiness} & Latent & Semantic & Boosted \\ \hline
        \multirow{4}{*}{MNIST}   & pHash   & 19.37  & 12.92    & 10.82   \\ \cline{2-5} 
        & MSE     & 0.23   & 0.03     & 0.01    \\ \cline{2-5} 
        & SSIM    & 3.81\%   & 77.14\%    & 88.32\%   \\ \cline{2-5} 
        & DLS     & 7.01   & 0.015    & 0.016   \\ \hline
    \end{tabular}%
\end{table}

\textcolor{black}{
\textbf{Robustness against defenses. }
We also investigate the stealthiness and robustness of the adversarial examples generated using the one-instance-based universal perturbation. The stealthiness of the results is shown in Table II. 
We can see that the perturbed instances using the perturbation derived from our semantic latent attack have higher SSIM similarity scores while smaller pHash, MSE, and DLS based dissimilarity scores, compared with using the ordinary latent attack. The results reveal that the universal perturbation using our attack can generate better quality adversarial examples in both input space and latent space to even fool a human. Namely, the stealthiness of the universal perturbation is better ensured by our attack.
We also find that the image with $boosted$ strategy can improve the reconstruction quality in both input space and latent space. 
\newline
We confirm the robustness against pixel-wise reconstruction based defense based on the MSE evaluation. The MSE values between the original and the reconstructed adversarial samples from our universal semantic adversarial attack are small enough to render them indistinguishable from clean samples, thus enabling the escape from detection. 
Additionally, DSL evaluations indicate a stronger resistance to adaptive latent variation defenses when compared to ordinary latent perturbation defenses. The DSL values of our semantic adversarial examples are also small enough to be indistinguishable from clean samples. 
\newline
Fig. 10 shows a visual demonstration of the perturbed images to further demonstrate the stealthiness of the universal perturbation. 
Compared with the universal perturbation for ordinary latent perturbation, the universal perturbation discovered using our semantic attack is more imperceptible and stealthy. 
We also find that such universal perturbations are not unique, as many different efficient universal perturbations can be generated for the same value range on a given latent factor. 
\newline
These experimental results show the existence of single universal perturbation, in terms of a specific latent factor, that causes clean images to be misclassified with high probability for one class of the instance. Such universal perturbations are also stealthy in both the latent and input space. 
}
\begin{figure}[!htb]
    \centering
    \setlength{\abovecaptionskip}{-0.05cm}
    \setlength{\belowcaptionskip}{-0.2cm}
    \includegraphics[width=3in,height=1.6in]{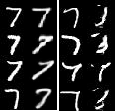}
    \caption{Visual demonstration of stealthiness for universal perturbation. Left subfigure is the misclassified image after adding universal perturbation into a latent factor that affects angles using a semantic attack, and the right subfigure is the misclassified image after adding universal perturbation into the latent vector using the ordinary latent attack.}
\end{figure}

\section{Related Work}
\textcolor{black}{
In terms of the adversarial investigation scope, we divide adversarial attacks into three categories, namely, nonstructural (pixel-level), semi-structural (layer-level), and structural (attribute-level). 
\newline
\textbf{Nonstructural level perturbations (pixel-level).} 
The most commonly adopted perturbation for adversarial examples is the artificially crafted pixel noise. However, they are nonstructural in the pixel-wise input space. 
Researchers developed several methods for generating such pixel adversarial perturbations, most of which leveraged gradient-based optimization from normal examples \cite{szegedy2013intriguing,carlini2017towards}. 
The fast gradient sign method (FGSM) \cite{goodfellow6572explaining} solves this optimization problem by performing a one-step gradient update from x in the image space with volume $\epsilon $. The update step-width $\epsilon $ is identical for each pixel, and the update direction is determined by the sign of gradient at this pixel. 
The projected gradient descent (PGD) attack \cite{madry2017towards} applies an iterative technique in which the gradient is projected rather than clipped in each step. 
DeepFool is also an iterative attack but formalizes the problem differently \cite{moosavi2016deepfool}. The basic idea is to find the closest decision boundary from a normal image x in the image space, and then to cross that boundary to fool the classifier.
Carlini recently introduced a powerful attack that generates adversarial examples with small perturbation \cite{carlini2017towards}. The attack can be targeted or untargeted for all three metrics $L^{0 }$, $L^{2 }$, and $L^{\infty}$. 
Moosavi et al. showed that it was even possible to find one effective universal adversarial perturbation that, when applied, turned many images adversarial \cite{moosavi2017universal}. 
As a consequence of these norm-bounded pixel perturbations, the adversarial examples are somewhat distinguishable from natural images. 
Multiple defense methods utilize this property. One strong defense is to detect or purify input data that may have added adversarial perturbation with hand-crafted statistical features \cite{grosse2017statistical}, separate classification networks \cite{metzen2017detecting} or detector-based denoising  \cite{meng2017magnet}. In MagNet \cite{meng2017magnet}, one or more separate detector networks and a reformer network are used to defend adversarial examples. 
The detector networks learn to differentiate between normal and adversarial examples by approximating the manifold of normal examples. The reformer network passes input data to the autoencoders that move input data closer to the data manifold. 
\newline
\textbf{Semi-structural level perturbations in pixel and latent space.} 
Due to the difficulty in interpreting nonstructural perturbations, new adversarial examples with structural perturbations are also developed in existing studies in order to provide some insight. 
To manipulate an image in pixel space or latent space, two types of transformation are generally applied. 
Common pixel transformations include rotation, clipping, or RGB conversion \cite{hosseini2018semantic,bhattad2019unrestricted,engstrom2018rotation}, which are based on the shape bias property of human perception. 
Semi-structural perturbations are always based on layer-level attributes, such as color, rotation, or portions, in order to produce photorealistic adversarial examples. However, the degree of granularity in such transformations is determined by the layer properties of the image, which are static, limited, and are not always applicable to all images. 
For the latent transformations, 
\cite{zhao2017generating} uses GAN to generate natural adversarial examples on the data manifold by searching in the semantic space of dense and continuous data representation. Specifically, it is designed to address the mismatch between the input space perturbation and the semantic space features. 
However, the perturbations in the latent space are difficult to control due to a lack of interpretability and factorization/disentanglement. 
In our knowledge, this is the first work that generates semantic adversarial examples based on the structural perturbation by manipulating the disentangled latent codes in terms of fine-granularity attributes about object.
}

\section{Conclusion and Discussion}
We propose a practical semantic adversarial attack with structured perturbations that can be conducted in a interpretable manner, by leveraging the disentangled representation. 
As compared to existing adversarial attack approaches, our proposed method offers three advantages.
As opposed to the existing adversarial examples, our adversarial perturbation is generalized to the wider object level. 
Unlike unstructured pixel-level or semi-structural layer-level adversarial perturbations, we propose a fine-granularity attribute-level adversarial perturbation. 
The perturbation design of our attack is lightweight, interpretable, and targets black-box classifiers for images as well. 

Our detailed experiments demonstrate that our attack schemes are both stealthy and efficient to generate adversarial examples in both a visual perspective and latent space. We also demonstrate that universal semantic adversarial examples exist in a significant number of samples with the same labels.

\textcolor{black}{
\textbf{Single domain and cross-domains attributes.} 
Due to the limitations of current generative models in terms of expressive power, the learned fine-granularity latent space representations tend to be restricted to the same domain of data.
The cross-objects semantic attributes in the same domain are investigated in this work. For example, the font size is the cross-object attribute among different digit objects (0-9). Our goal is to develop a structural perturbation or semantic perturbation that simulates the performance of the trained classifier when faced with variations of objects' attributes in the real world. Furthermore, one of our future research is to quantify the robustness of the model with respect to the fine-granularity attributes. 
\newline
Unlike the existing adversarial examples specified for a single sample level, our adversarial perturbation is generalized for the broader object level. 
Besides, we defined the universal semantic adversarial example (Section 3.1) to evaluate the transferability among samples that belong to the same object or different objects. 
Our results in the Section 5.5.3 demonstrate such transferability in detail. 
\newline
It is one of the frontier research topics in the deep generative model field to learn more expressive and comprehensive feature or attribute representation for cross-domain data. 
It is possible to find some cross-dataset or cross-domain semantic attributes in the latent space, such as the latent code that affects the background of the images.
However, the granularity of such semantic attributes is more coarse and not relevant to the object. A potential refinement of our work includes designing universal perturbation for multiple domain data. We also aim to investigate applying variants of the semantic attack to other domains besides image classification, e.g., time-series data and text data. 
}

\textcolor{black}{
\textbf{VAE and GAN. }
As input, GANs typically generate samples from random noise, and then the samples are evaluated and differentiated using a discriminator. However, this additional encoder must be trained to reverse an image into its latent representation in the latent space of the generator, i.e., image reverse encoding. For the sake of simplicity, we consider VAE to be the generative scheme, with the discriminator of GAN to enhance the reconstruction quality. 
On the other hand, the latest GAN variants, such as StyleGAN, have demonstrated outstanding visual quality and fidelity for the generation of images using latent spaces with style and content properties. However, such disentanglement is often abstractive level information, such as texture information. The purpose of this work is to implement a finer granularity level of disentanglement with respect to various attributes of the object. It is still an open question to harness the latent space of complicated GANs into finer-granularity disentanglement with less supervision. Our following research endeavors include developing a reverse encoding method to encode a given image into the latent representations of state-of-the-art GANs, and refining the granularity of the latent representation to include attributes instead of only the abstract level. Another refinement of our work includes further addressing the disentanglement and reconstruction trade-off.
}

\textcolor{black}{
\textbf{The limitations of generative models.} 
In general, based on the representation capability of existing generative models, the fine-granularity attribute level disentangled latent representation could be obtained only for a single object. 
Therefore, the initial step of our work is to exploit the fragility of the model in terms of more fine-granularity attributes specified for single object scenario, such as digit classification and face recognition. 
Multiple objects are also involved in many tasks, such as object detection, which not only performs image classification but also determines the positions of objects within an image. Before we proceed to exploit the effects of our semantic adversarial perturbation on object detection tasks, we must address the open challenge for a generative model on multiple objection representation. 
For multi-object generative models, the main bottleneck is the need for a sufficient number of training samples for each object.
To alleviate the restriction of available data, one solution is data augmentation, such as Differentiable Augmentation (DiffAugment) \cite{zhao2020differentiable}, where the performance of the model trained using only 100 images for the augmentation could match that of the model trained using the entire dataset.
Our next step will be to use data augmentation and transfer learning to address the issue of data limitations.  
Based on the representation of multiple objects, our further step is to exploit the performance of our attacks on the object detection tasks.
}
\bibliographystyle{IEEEtran}
\bibliography{tnnls}
\end{document}